\documentclass[nohyperref]{article}

\usepackage{microtype}
\usepackage{graphicx}
\usepackage{booktabs} 

\usepackage{hyperref}
\usepackage[normalem]{ulem}

\usepackage[accepted]{icml2022}

\usepackage{amsmath}
\usepackage{amssymb}
\usepackage{mathtools}
\usepackage{amsthm}

\usepackage[capitalize,noabbrev]{cleveref}

\theoremstyle{plain}

\theoremstyle{definition}

\theoremstyle{remark}

\usepackage[textsize=tiny]{todonotes}

\usepackage{microtype}

\usepackage{multirow}
\usepackage{graphics}
\usepackage{amsmath}
\usepackage{multicol}
\usepackage{lipsum}
\usepackage{mwe}
\usepackage{hyperref}
\usepackage{url}
\usepackage[caption=true]{subfig}
\usepackage{xcolor}
\usepackage{color}
\usepackage{tcolorbox}
\usepackage{bbm}
\usepackage{upgreek}

\usepackage{wrapfig}
\usepackage{colortbl}

\usepackage{amsmath,amsfonts,bm}

\def\eqref#1{equation~\ref{#1}}

\def\ceil#1{\lceil #1 \rceil}

\def\1{\bm{1}}

\DeclareMathAlphabet{\mathsfit}{\encodingdefault}{\sfdefault}{m}{sl}
\SetMathAlphabet{\mathsfit}{bold}{\encodingdefault}{\sfdefault}{bx}{n}

\newcommand{\R}{\mathbb{R}}

\DeclareMathOperator*{\argmax}{arg\,max}

\newcommand{\method}{\textsc{UNIREX}}
\newcommand{\methodsp}{\textsc{UNIREX}\xspace}
\newcommand{\eg}{\textit{e.g., }}
\newcommand{\ie}{\textit{i.e., }}

\colorlet{lred}{red!15}
\colorlet{lblue}{blue!15}
\colorlet{lgreen}{green!20}

\usepackage[symbol]{footmisc}

\DeclareUnicodeCharacter{FFFD}{ }

\icmltitlerunning{\method: A Unified Learning Framework for Language Model Rationale Extraction}

\begin{document}

\twocolumn[
\icmltitle{\method: A Unified Learning Framework for \\ Language Model Rationale Extraction}




\begin{icmlauthorlist}
\icmlauthor{Aaron Chan}{usc}
\icmlauthor{Maziar Sanjabi}{fb}
\icmlauthor{Lambert Mathias}{fb}
\icmlauthor{Liang Tan}{fb} \\
\icmlauthor{Shaoliang Nie}{fb}
\icmlauthor{Xiaochang Peng}{fb}
\icmlauthor{Xiang Ren}{usc}
\icmlauthor{Hamed Firooz}{fb}
\end{icmlauthorlist}

\icmlaffiliation{usc}{University of Southern California}
\icmlaffiliation{fb}{Meta AI}

\icmlcorrespondingauthor{Aaron Chan}{aarzchan@gmail.com}

\icmlkeywords{explainability, interpretability, transparency, rationale extraction, faithfulness, plausibility, language model, text classification, machine learning, deep learning, natural language processing}

\vskip 0.3in
]



\printAffiliationsAndNotice{}  

\begin{abstract}



An extractive rationale explains a language model's (LM's) prediction on a given task instance by highlighting the text inputs that most influenced the prediction.
Ideally, rationale extraction should be \emph{faithful} (reflective of LM's actual behavior) and \emph{plausible} (convincing to humans), without compromising the LM's (\ie task model's) \emph{task performance}.
Although attribution algorithms and select-predict pipelines are commonly used in rationale extraction, they both rely on certain heuristics that hinder them from satisfying all three desiderata.
In light of this, we propose \method, a flexible learning framework that generalizes rationale extractor optimization as follows: (1) specify architecture for a learned rationale extractor; (2) select explainability objectives (\ie faithfulness and plausibility criteria); and (3) jointly train the task model and rationale extractor on the task using the selected objectives.
\methodsp enables replacing prior works' heuristic design choices with a generic learned rationale extractor in (1) and optimizing it for all three desiderata in (2)-(3).
To facilitate comparison between methods with respect to multiple desiderata, we introduce the Normalized Relative Gain (NRG) metric.
On five English text classification datasets, our best \methodsp configuration outperforms baselines by an average of 32.9\% NRG.
Plus, \methodsp rationale extractors' faithfulness can even generalize to unseen datasets and tasks.\footnote[2]{Code: \href{https://github.com/facebookresearch/UNIREX}{github.com/facebookresearch/UNIREX}.}

\end{abstract}



\section{Introduction} 
\label{sec:intro}

\begin{figure}[t!]
\centering
\includegraphics[width=0.48\textwidth]{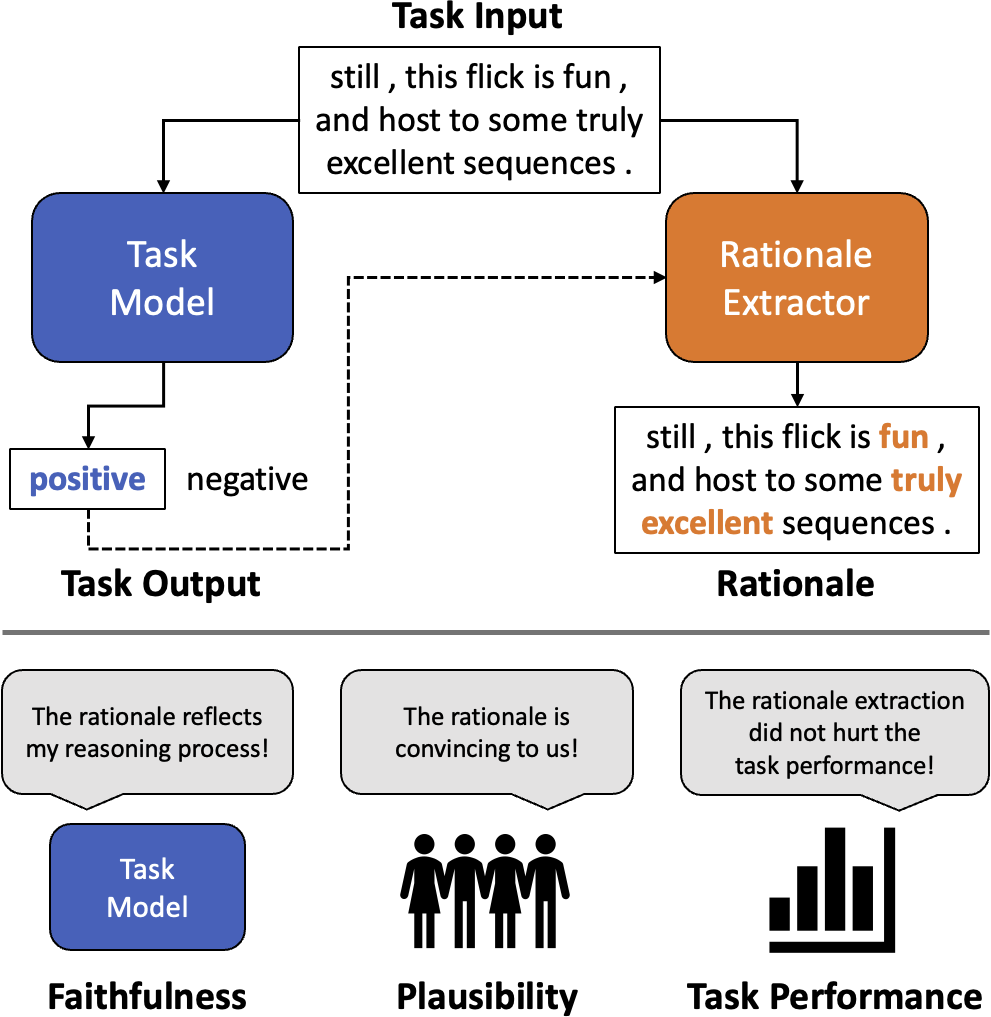}
\caption{\small \textbf{Desiderata of Rationale Extraction.}
Ideally, rationale extraction should be \textit{faithful} and \textit{plausible}, without compromising the task model's \textit{task performance}.
Unlike prior works, \methodsp enables optimizing the rationale extractor for all three desiderata.
}
\vspace{-0.4cm}
\label{fig:rationale}
\end{figure}

In recent years, neural language models (LMs) have yielded state-of-the-art performance on a wide range of natural language processing (NLP) tasks \citep{devlin2018bert,liu2019roberta}.
However, LMs' complex processes are notoriously opaque \citep{rudin2019stop}, posing concerns about the societal implications of using LMs for high-stakes decision-making \citep{bender2021dangers}.
Thus, explaining LMs' behavior is crucial for promoting trust, ethics, and safety in NLP systems \citep{doshi2017towards, lipton2018mythos}.
Given a LM's (\ie task model's) predicted label on a text classification instance, an \textit{extractive rationale} is a type of explanation that highlights the tokens that most influenced the model to predict that label \citep{luo2021local}.
To provide meaningful explanations, rationale extraction should be \emph{faithful} (reflective of LM's actual behavior) \citep{ismail2021improving, jain2020learning} and \emph{plausible} (convincing to humans) \citep{deyoung2019eraser}, without compromising the LM's \emph{task performance} \citep{deyoung2019eraser, jacovi2020towards} (Fig. \ref{fig:rationale}).

Configuring the rationale extractor and its training process can greatly impact these desiderata, yet prior works have commonly adopted at least one of the following suboptimal heuristic design choices.
First, many works rely in some way on \textit{attribution algorithms} (AAs), which extract rationales via handcrafted functions \citep{sundararajan2017axiomatic, ismail2021improving, situ2021learning}.
AAs may have built-in faithfulness-related properties but cannot be directly trained and tend to be compute-intensive \citep{bastings2020elephant}.
The most similar work to ours is SGT \citep{ismail2021improving}, which regularizes a task model to produce faithful AA-based rationales.
Still, AAs can be a bottleneck for plausibility, as producing human-like rationales is a complex objective requiring high capacity rationale extractors \citep{narang2020wt5, deyoung2019eraser}.
Second, many works use a specialized \textit{select-predict pipeline} (SPP), where a predictor module is trained to solve the task using only tokens chosen by a selector module \citep{jain2020learning, yu2021understanding, paranjape2020information}.
Instead of faithfulness optimization, SPPs heuristically aim for ``faithfulness by construction" by treating the selected tokens as a rationale for the predictor's output (which depends only on those tokens).
Still, SPPs typically have worse task performance than vanilla LMs since SPPs hide the full input from the predictor and are hard to train end-to-end \citep{jain2020learning, bastings2019interpretable, lei2016rationalizing}.
Both AAs and SPPs utilize heuristics that fundamentally limit the rationale extractor from achieving all three desiderata.

To tackle this challenge, we propose the \textbf{UNI}fied Learning Framework for \textbf{R}ationale \textbf{EX}traction (\textbf{\method}), which generalizes rationale extractor optimization as follows: (1) specify architecture for a learned rationale extractor; (2) select explainability objectives (\ie faithfulness and plausibility criteria); and (3) jointly train the task model and rationale extractor on the task using selected objectives (\textsection \ref{sec:method}).
\methodsp enables replacing prior works' heuristic design choices in (1) with a generic learned rationale extractor and optimizing it for all three desiderata in (2)-(3).

\methodsp provides great flexibility in performing (1)-(3).
For (1), any model architecture is applicable, but we study Transformer LM based rationale extractors in this work \citep{zaheer2020big, deyoung2019eraser}.
We focus on two architectures: (A) Dual LM, where task model and rationale extractor are separate; and (B) Shared LM, where task model and rationale extractor share parameters.
For (2), any faithfulness and plausibility criteria can be used.
Following \citet{deyoung2019eraser}, we focus on comprehensiveness and sufficiency as faithfulness criteria, while using similarity to gold rationales as plausibility criteria.
For (3), trade-offs between the three desiderata can be easily managed during rationale extractor optimization by setting arbitrary loss weights for the faithfulness and plausibility objectives.
Furthermore, although computing the faithfulness criteria involves discrete (non-differentiable) token selection, using the Shared LM architecture can approximate end-to-end training and enable both task model and rationale extractor to be optimized with respect to all three desiderata (\textsection \ref{sec:method:training}).


To evaluate all three desiderata in aggregate, we introduce the Normalized Relative Gain (NRG) metric.
On five English text classification datasets -- SST, Movies, CoS-E, MultiRC, and e-SNLI \citep{carton2020evaluating, deyoung2019eraser} -- our best \methodsp configuration outperforms the strongest baselines by an average of 32.9\% NRG (\textsection \ref{sec:experiments:main}), showing that \methodsp can optimize rationale extractors for all three desiderata.
In addition, we verify our \methodsp design choices via extensive ablation studies (\textsection \ref{sec:experiments:ablation}).
Moreover, \method-trained extractors have considerable generalization power, yielding high plausiblity with minimal gold rationale supervision (\textsection \ref{sec:experiments:gold_eff}) and high faithfulness on unseen datasets/tasks (\textsection \ref{sec:experiments:zs}).
Finally, our user study shows that humans judge \methodsp rationales as more plausible than rationales extracted via other methods (\textsection \ref{sec:experiments:user_study}).

\section{Problem Formulation} 
\label{sec:background}

We formalize rationale extraction and discuss how extractive rationales are evaluated, in the context of text classification.


\subsection{Rationale Extraction}
\label{sec:background:rationale_extr}

Here, we consider $\mathcal{F}_{\text{task}} = f_{\text{task}}(f_{\text{enc}}(\cdot))$ as a task model for $M$-class text classification (\textsection \ref{sec:appendix:text_cls}), where $f_{\text{enc}}$ is the text encoder while $f_{\text{task}}$ is the task output head.
In modern NLP systems, $\mathcal{F}_{\text{task}}$ usually has a BERT-style architecture \citep{devlin2018bert}, in which $f_{\text{enc}}$ is a Transformer network \citep{vaswani2017attention} while $f_{\text{task}}$ is a linear layer with softmax classifier.
Let $\mathbf{x}_i = [x_{i}^{t}]_{t=1}^{n}$ be the $n$-token input sequence (\eg a sentence) for task instance $i$, and $\mathcal{F}_{\text{task}}(\mathbf{x}_i) \in \R^{M}$ be the logit vector for the output of the task model.
We use $y_i = \argmax_{\hspace{0.5mm} j} \mathcal{F}_{\text{task}}(\mathbf{x}_i)_j$ to denote the class predicted by $\mathcal{F}_{\text{task}}$.
Given $\mathcal{F}_{\text{task}}$, $\mathbf{x}_i$, and $y_i$, the goal of rationale extraction is to output vector $\mathbf{s}_i = [s_{i}^{t}]_{t=1}^{n}\in \R^{n}$, such that each $s_{i}^{t} \in \mathbb{R}$ is an \textit{importance score} indicating how strongly token $x_{i}^{t}$ influenced $\mathcal{F}_{\text{task}}$ to predict class $y_i$.

Let $\mathcal{F}_{\text{ext}}$ denote a rationale extractor, such that $\mathbf{s}_i = \mathcal{F}_{\text{ext}}(\mathcal{F}_{\text{task}}, \mathbf{x}_i, y_i)$.
$\mathcal{F}_{\text{ext}}$ can be a learned model or heuristic function.
In practice, the final rationale is typically obtained by binarizing $\mathbf{s}_i$ as $\mathbf{r}_i \in \{0, 1\}^{n}$, via the top-$k\%$ strategy: $r_{i}^{t} = 1$ if $s_{i}^{t}$ is one of the top-$k\%$ scores in $\mathbf{s}_i$; otherwise, $r_{i}^{t} = 0$ \citep{deyoung2019eraser, jain2020learning, pruthi2020evaluating, chan2021salkg}.
While other binarization strategies can be used (\eg score threshold, highest-scoring contiguous $k$-token span), we focus on top-$k\%$ in this study, since this strategy is most prevalent in the explainability literature.
For top-$k\%$, let $\mathbf{r}_{i}^{(k)}$ denote the ``important" (\ie ones) tokens in $\mathbf{r}_{i}$, when using $0 \leq k \leq 100$. 


\subsection{Three Desiderata of Rationale Extraction}
\label{sec:background:desiderata}




To provide meaningful explanations, rationale extraction via $\mathcal{F}_{\text{ext}}$ should be \textit{faithful} and \textit{plausible}, without significantly hurting $\mathcal{F}_{\text{task}}$'s \textit{task performance} \citep{deyoung2019eraser}.

\paragraph{Faithfulness} 
Faithfulness means how accurately a rationale reflects $\mathcal{F}_{\text{task}}$'s true reasoning process for predicting $y_i$ \citep{jacovi2020towards}.
Hence, faithfulness metrics aim to measure the extent to which the $\mathbf{r}_{i}^{(k)}$ tokens influence $p_{y_i}(\mathbf{x}_i)$, which denotes $\mathcal{F}_{\text{task}}$'s confidence probability for $y_i$ when using $\mathbf{x}_i$ as input \citep{deyoung2019eraser, shrikumar2017learning, hooker2018benchmark, pruthi2020evaluating}.
Recently, comprehensiveness and sufficiency have emerged as popular faithfulness metrics in the explainability literature \citep{deyoung2019eraser}.
\textbf{\textit{Comprehensiveness}} (comp) measures the change in $p_{y_i}$ when $\mathbf{r}_{i}^{(k)}$ is \textit{removed} from the input: $\text{comp} = p_{y_i}(\mathbf{x}_i) - p_{y_i}(\mathbf{x}_i \backslash \mathbf{r}_{i}^{(k)})$.
That is, if the $\mathbf{r}_{i}^{(k)}$ tokens are truly influential, then removing them from the input should decrease $\mathcal{F}_{\text{task}}$'s predicted probability for $y_i$.
Thus, higher comp indicates higher faithfulness.
\textbf{\textit{Sufficiency}} (suff) measures the change in $p_{y_i}$ when only $\mathbf{r}_{i}^{(k)}$ is \textit{kept} in the input: $\text{suff} = p_{y_i}(\mathbf{x}_i) - p_{y_i}(\mathbf{r}_{i}^{(k)})$.
That is, if the $\mathbf{r}_{i}^{(k)}$ tokens are truly influential, only keeping them in the input should not decrease $\mathcal{F}_{\text{task}}$'s predicted probability for $y_i$.
Thus, lower suff indicates higher faithfulness.

\paragraph{Plausibility} 
Plausibility is defined as how convincingly a rationale explains a given model's prediction, as judged by humans \citep{jacovi2020towards}. 
This can be measured either by automatically computing the similarity between $\mathcal{F}_{\text{ext}}$'s rationales (either $\mathbf{s}_i$ or $\mathbf{r}_i$) and human-annotated gold rationales \citep{deyoung2019eraser}, or by asking human annotators to rate whether $\mathcal{F}_{\text{ext}}$'s rationales make sense for predicting $y_i$ \citep{strout2019human, doshi2017towards}.
Typically, a gold rationale is a binary vector $\mathbf{r}_{i}^{*} \in \{0, 1\}^{n}$, where ones and zeros indicate important and unimportant tokens, respectively \citep{lei2016rationalizing, deyoung2019eraser}.


 
\paragraph{Task Performance}
Task performance, in the context of rationale extraction, concerns how much $\mathcal{F}_{\text{task}}$'s task performance (on the test set) drops when $\mathcal{F}_{\text{task}}$ is trained with explainability objectives (\ie faithfulness, plausibility) for $\mathcal{F}_{\text{ext}}$.
As long as $\mathcal{F}_{\text{task}}$ is trained with non-task losses, $\mathcal{F}_{\text{task}}$'s task performance can be affected.
Note that this means post hoc (\ie introduced after $\mathcal{F}_{\text{task}}$ training is over) rationale extraction will not affect $\mathcal{F}_{\text{task}}$'s task performance.
In general, the main goal of $\mathcal{F}_{\text{task}}$ is high task performance, so we should ideally improve $\mathcal{F}_{\text{ext}}$ with respect to the other desiderata without hurting $\mathcal{F}_{\text{task}}$'s task performance.
To measure task performance, we use standard dataset-specific performance metrics (\eg accuracy, F1).


\section{\method} 
\label{sec:method}


\begin{figure*}[t]
\centering
\includegraphics[width=\textwidth]{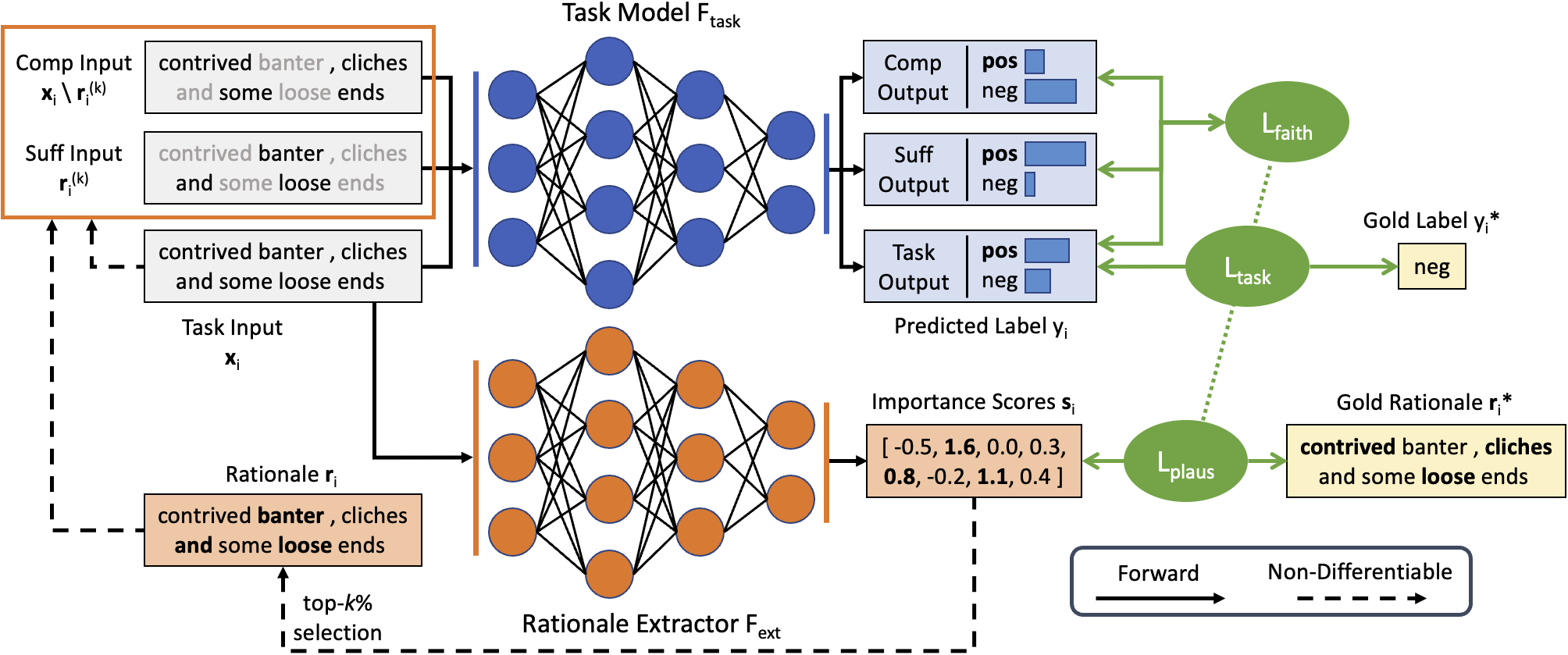}
\caption{\small \textbf{\methodsp Framework.} \methodsp enables us to jointly optimize the task model ($\mathcal{F}_{\text{task}}$) and rationale extractor ($\mathcal{F}_{\text{ext}}$), with respect to faithfulness ($\mathcal{L}_{\text{faith}}$), plausibility ($\mathcal{L}_{\text{plaus}}$), and task performance ($\mathcal{L}_{\text{task}}$).
In this example, we consider the sentiment analysis task.
For \textbf{task performance}, $\mathcal{F}_{\text{task}}$ is trained via gold label $y_{i}^{*}$ to predict the sentiment -- either positive (pos) or negative (neg) -- of sentence $\mathbf{x}_i$.
Here, $\mathcal{F}_{\text{task}}$'s predicted label for $\mathbf{x}_i$ is $y_i$ = pos.
For \textbf{plausibility}, $\mathcal{F}_{\text{ext}}$ is trained via gold rationale $\mathbf{r}^{*}_i$ to output human-aligned token importance scores $\mathbf{s}_i$ for $\mathbf{x}_i$ (\textsection \ref{sec:method:expl_objectives:plausibility}).
For \textbf{faithfulness}, $\mathbf{s}_i$ is binarized as rationale $\mathbf{r}_i$ via top-$k$\% selection, then used to construct the comp ($\mathbf{x}_i \backslash \mathbf{r}_{i}^{(k)}$) and suff ($\mathbf{r}_{i}^{(k)}$) inputs for $\mathcal{L}_{\text{task}}$. 
With $\mathcal{L}_{\text{task}}$'s predicted probabilities for $y_i$, given $\mathbf{x}_i$, $\mathbf{x}_i \backslash \mathbf{r}_{i}^{(k)}$, and $\mathbf{r}_{i}^{(k)}$, respectively, the comp and suff losses are computed.
The comp and suff losses align $\mathcal{L}_{\text{task}}$'s output with $\mathbf{r}_i$, such that $\mathbf{r}_i$ becomes a faithful explanation of $\mathcal{L}_{\text{task}}$'s behavior. (\textsection \ref{sec:method:expl_objectives:faithfulness}).
Note that some parts of \methodsp are non-differentiable.
Still, by having $\mathcal{L}_{\text{task}}$ and $\mathcal{L}_{\text{ext}}$ share a text encoder, we can approximate end-to-end training of both models, jointly with respect to all three desiderata (\textsection \ref{sec:method:training}).
}
\label{fig:unirex}
\end{figure*}

We present the \methodsp learning framework, which enables us to jointly optimize the task model and rationale extractor with respect to faithfulness, plausibility, and task performance.

\subsection{Framework Overview}
\label{sec:method:overview}


Given task model $\mathcal{F}_{\text{task}}$, \methodsp generalizes rationale extractor optimization as follows: (1) choose architecture for a learned rationale extractor $\mathcal{F}_{\text{ext}}$; (2) select explainability objectives (\ie faithfulness loss $\mathcal{L}_{\text{faith}}$ and plausibility loss $\mathcal{L}_{\text{plaus}}$); and (3) jointly train $\mathcal{F}_{\text{task}}$ and $\mathcal{F}_{\text{ext}}$ using $\mathcal{L}_{\text{task}}$ (task loss), $\mathcal{L}_{\text{faith}}$, and $\mathcal{L}_{\text{plaus}}$.
As shown in Fig. \ref{fig:unirex}, \methodsp training consists of two backpropagation paths.
The first path is used to update $\mathcal{F}_{\text{task}}$ with respect to $\mathcal{L}_{\text{task}}$ and $\mathcal{L}_{\text{faith}}$.
Whereas $\mathcal{L}_{\text{task}}$ is computed with respect to the task target $y_{i}^{*}$, $\mathcal{L}_{\text{faith}}$ is computed only using the task input $\mathbf{x}_{i}$ and the top-$k\%$ important tokens $\mathbf{r}_{i}^{(k)}$ (obtained via $\mathcal{F}_{\text{ext}}$), based on some combination of comp and suff (\textsection \ref{sec:background:desiderata}).
The second path is used to update $\mathcal{F}_{\text{ext}}$ with respect to $\mathcal{L}_{\text{plaus}}$, which encourages importance scores $\mathbf{s}_{i}$ to approximate gold rationale $\mathbf{r}_{i}^{*}$.

Thus, \methodsp frames rationale extraction as the following optimization problem:
\vspace{-5mm}

\begin{equation}
\begin{split}
\min_{\mathcal{F}_{\text{task}}, \hspace{0.5mm} \mathcal{F}_{\text{ext}}}  &\mathcal{L}_{\text{task}}(\mathbf{x}_i, y_{i}^{*}; \mathcal{F}_{\text{task}}) \\
&+ \hspace{1mm} \alpha_{f} \mathcal{L}_{\text{faith}}(\mathbf{x}_i, \mathbf{r}_{i}^{(k)}; \mathcal{F}_{\text{task}}) \\
&+ \hspace{1mm} \alpha_{p} \mathcal{L}_{\text{plaus}}(\mathbf{x}_i, \mathbf{r}_{i}^{*}; \mathcal{F}_{\text{ext}}),
\end{split}
\end{equation}

where $\alpha_{f}$ and $\alpha_{p}$ are loss weights.
If $\mathcal{F}_{\text{task}}$ and $\mathcal{F}_{\text{ext}}$ share parameters, then the shared parameters will be optimized with respect to all losses.
During inference, for task input $\mathbf{x}_i$, we first use $\mathcal{F}_{\text{task}}$ to predict $y_{i}^{*}$, then use $\mathcal{F}_{\text{ext}}$ to output a rationale $\mathbf{r}_{i}$ for $\mathcal{F}_{\text{task}}$'s prediction $y_i$.
Below, we discuss different options for \methodsp's rationale extractor and explainability objectives.




\begin{figure*}[ht]
	\centering
	\vspace{0.4cm}
	\includegraphics[width=0.8\textwidth]{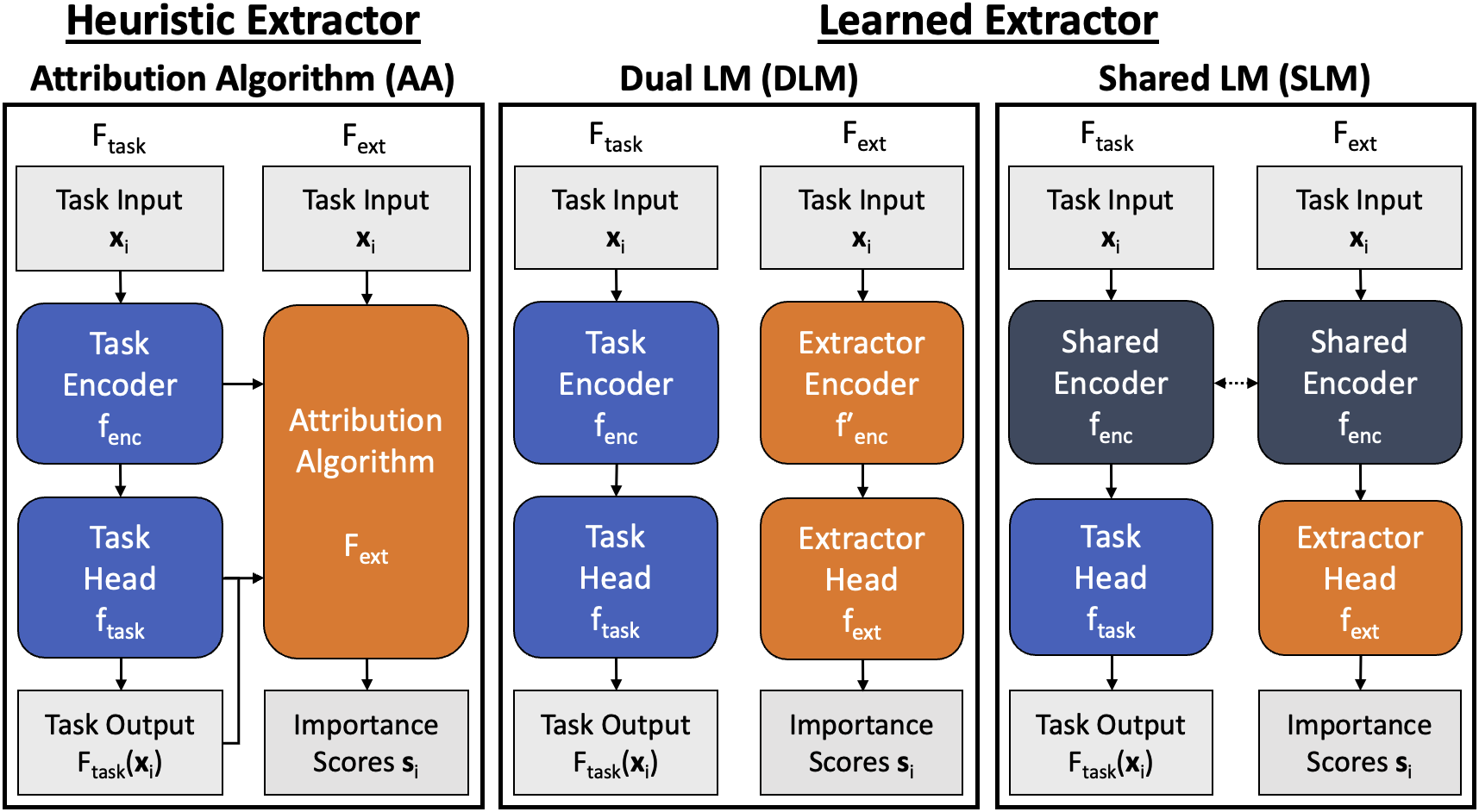}
	\caption{\small \textbf{Rationale Extractor Types.}
		In general, rationale extractor $\mathcal{L}_{\text{ext}}$ can be either heuristic or learned.
		A heuristic $\mathcal{L}_{\text{ext}}$ is a handcrafted \textbf{attribution algorithm}, which cannot be trained (\textsection \ref{sec:method:extractor:heuristic}).
		By default, \methodsp uses a learned $\mathcal{L}_{\text{ext}}$, which can be optimized for faithfulness, plausibility, and task performance.
		For learned $\mathcal{L}_{\text{ext}}$, we focus on two architectures (w.r.t. task model $\mathcal{L}_{\text{task}}$): \textbf{Dual LM} designs $\mathcal{L}_{\text{task}}$ and $\mathcal{L}_{\text{ext}}$ as two fully separate LMs, while \textbf{Shared LM} designs $\mathcal{L}_{\text{task}}$ and $\mathcal{L}_{\text{ext}}$ to share the same text encoder (\textsection \ref{sec:method:extractor:learned}).
		Although some operations within \methodsp are non-differentiable, Shared LM's shared encoder allows us to approximate end-to-end training of both models w.r.t. all three desiderata (\textsection \ref{sec:method:training}, Fig. \ref{fig:unirex}).
	}
	\label{fig:extractor_types}
\end{figure*}

\subsection{Rationale Extractor}
\label{sec:method:extractor}

In \method, $\mathcal{F}_{\text{ext}}$ is a learned function by default.
Here, we first introduce heuristic $\mathcal{F}_{\text{ext}}$ (\ie AA), then discuss why a learned $\mathcal{F}_{\text{ext}}$ should typically be preferred (\textsection \ref{sec:background:rationale_extr}).
For each $\mathcal{F}_{\text{ext}}$ type, we present several possible design choices and the pros/cons of the given type.

\subsubsection{Heuristic Rationale Extractors}
\label{sec:method:extractor:heuristic}
Heuristic $\mathcal{F}_{\text{ext}}$ refers to AAs, which can be any handcrafted function that calculates an importance score $s_{i}^{t}$ for each input token $x_{i}^{t}$ \citep{bastings2020elephant}.
AAs are typically gradient-based \citep{sundararajan2017axiomatic, denil2014extraction, lundberg2017unified, li2015visualizing} or perturbation-based \citep{li2016understanding, poerner2018evaluating, kadar2017representation} methods.
Recall that $p_{y_i}(\mathbf{x}_i)$ denotes $\mathcal{F}_{\text{task}}$'s predicted probability for class $y_i$ (\textsection \ref{sec:background:desiderata}).
Gradient-based methods compute $s_{i}^{t}$ via the gradient of $p_{y_i}(\mathbf{x}_i)$ with respect to $x_{i}^{t}$. These methods require one or more $\mathcal{F}_{\text{task}}$ backward passes.
Perturbation-based methods measure $s_{i}^{t}$ as $p_{y_i}(\mathbf{x}_i)$'s change when perturbing (\eg removing) $x_{i}^{t}$.
These methods require multiple $\mathcal{F}_{\text{task}}$ forward passes -- typically, one forward pass per token in $\mathbf{x}_{i}$.

AAs can be used out of the box without training and are designed to satisfy certain faithfulness-related axiomatic properties \citep{sundararajan2017axiomatic, lundberg2017unified}.
However, AAs' lack of learnable parameters means they cannot be optimized for faithfulness/plausibility.
Thus, if $\mathcal{F}_{\text{task}}$ is trained for explainability using AA-based rationales, then only $\mathcal{F}_{\text{task}}$ is optimized.
Also, faithful AAs tend to be compute-intensive, requiring many $\mathcal{F}_{\text{task}}$ backward/forward passes per instance \citep{sundararajan2017axiomatic, lundberg2017unified, li2016understanding}.

\subsubsection{Learned Rationale Extractors}
\label{sec:method:extractor:learned}
Learned $\mathcal{F}_{\text{ext}}$ can be any learned model that transforms $x_{i}^{t}$ into $s_{i}^{t}$.
Given their success in NLP explainability \citep{deyoung2019eraser}, we focus on pre-trained Transformer LMs and highlight two key architectures: Dual LM (DLM) and Shared LM (SLM) (Fig. \ref{fig:extractor_types}).
For DLM, $\mathcal{F}_{\text{task}}$ and $\mathcal{F}_{\text{ext}}$ are two separate Transformer LMs with the same encoder architecture.
Formally, we define the DLM extractor as $\mathcal{F}_{\text{ext}} = f_{\text{ext}}(f_{\text{enc}}'(\cdot))$, where $f_{\text{enc}}'$ and $f_{\text{ext}}$ are $\mathcal{F}_{\text{ext}}$'s encoder and output head, respectively.
DLM provides more capacity for $\mathcal{F}_{\text{ext}}$, which can help $\mathcal{F}_{\text{ext}}$ output plausible rationales.
For SLM, $\mathcal{F}_{\text{task}}$ and $\mathcal{F}_{\text{ext}}$ are two Transformer LMs sharing encoder $f_{\text{enc}}$, while $\mathcal{F}_{\text{ext}}$ has its own output head $f_{\text{ext}}$.
Formally, the SLM extractor is defined as $\mathcal{F}_{\text{ext}} = f_{\text{ext}}(f_{\text{enc}}(\cdot))$.
SLM leverages multitask learning between $\mathcal{F}_{\text{task}}$ and $\mathcal{F}_{\text{ext}}$, which can improve faithfulness since $\mathcal{F}_{\text{ext}}$ has greater access to information about $\mathcal{F}_{\text{task}}$'s reasoning process.
By default, $\mathcal{F}_{\text{ext}}$ takes $\mathbf{x}_i$ as input and uses a linear layer for $f_{\text{ext}}$, although these settings can be changed if desired.

Unlike heuristic $\mathcal{F}_{\text{ext}}$, learned $\mathcal{F}_{\text{ext}}$ can be optimized for faithfulness/plausibility and only require one $\mathcal{F}_{\text{task}}$ forward pass during inference (\eg perturbation-based AAs require $n$ forward passes per $n$-token instance).
However, they cannot be used out of the box without explainability training and do not have built-in axiomatic properties -- \eg sensitivity and implementation invariance in the IG \citep{sundararajan2017axiomatic} AA -- which are designed to promote faithfulness.

Overall, learned $\mathcal{F}_{\text{ext}}$ is preferred if: (A) the goal is to optimize for both faithfulness and plausibility, and (B) gold rationales -- even a small amount -- are available for plausibility optimization. 
(B) is true because gold rationale annotated instances can be provided in every batch via oversampling (Sec \ref{sec:appendix:gold}), which works surprisingly well in low-resource settings (\textsection \ref{sec:experiments:gold_eff}).
Otherwise, \methodsp allows for the learned $\mathcal{F}_{\text{ext}}$ to be replaced with a heuristic $\mathcal{F}_{\text{ext}}$.




\subsection{Explainability Objectives}
\label{sec:method:expl_objectives}

After selecting $\mathcal{F}_{\text{ext}}$, we specify the explainability objectives, which can be any combination of faithfulness and plausibility criteria.
In prior approaches (\eg AA, SPPs), the rationale extractor is not optimized for both faithfulness and plausibility, but \methodsp makes this possible.
For any choice of learned $\mathcal{F}_{\text{ext}}$, \methodsp lets us easily ``plug and play'' different criteria and loss weights, based on our needs and domain knowledge, to find those that best balance the rationale extraction desiderata.

\subsubsection{Faithfulness}
\label{sec:method:expl_objectives:faithfulness}
Faithfulness refers to how accurately a rationale (output by $\mathcal{F}_{\text{ext}}$) reflects $\mathcal{F}_{\text{task}}$'s decision process for a given instance.
Evaluating rationale faithfulness is still an open problem with numerous applicable metrics, and \methodsp is not tailored for any specific metric.
However, given the prevalence of comp and suff (\textsection \ref{sec:background:rationale_extr}) in the explainability literature \citep{deyoung2019eraser, ismail2021improving}, we focus on comp and suff related objectives.

Recall that comp measures the importance of tokens in $\mathbf{r}_{i}^{(k)}$ as how $p_{y_i}(\mathbf{x}_i)$ changes when those tokens are removed from $\mathbf{x}_i$.
Intuitively, we want $p_{y_i}(\mathbf{x}_i)$ to be higher than $p_{y_i}(\mathbf{x}_i \backslash \mathbf{r}_{i}^{(k)})$, so higher comp is better.
Since comp is defined for a single class' probability rather than the label distribution, we can define the comp loss $\mathcal{L}_{\text{comp}}$ via cross-entropy loss $\mathcal{L}_\text{CE}$ (which is computed w.r.t. the target class), as in the following \textit{difference criterion} instantiation of $\mathcal{L}_{\text{comp}}$:
\begin{equation}
\begin{split}
\mathcal{L}_{\text{comp-diff}} = &\mathcal{L}_{\text{CE}}(\mathcal{F}_{\text{task}}(\mathbf{x}_i), y_{i}^{*}) \\
 &- \mathcal{L}_{\text{CE}}(\mathcal{F}_{\text{task}}(\mathbf{x}_i \backslash \mathbf{r}_{i}^{(k)}), y_{i}^{*})
\end{split}
\end{equation}

\begin{equation}
\mathcal{L}_\text{CE}(\mathcal{F}_{\text{task}}(\mathbf{x}_i), y_{i}^{*}) = -y_{i}^{*}\log(\mathcal{F}_{\text{task}}(\mathbf{x}_i))
\end{equation}

For training stability, we compute comp loss for target class $y_{i}^{*}$ here instead of $\mathcal{F}_{\text{task}}$'s predicted class $y_i$, since $y_i$ is a moving target during training.
Using $\mathcal{L}_{\text{comp-diff}}$, it is possible for $\mathcal{L}_{\text{CE}}(\mathcal{F}_{\text{task}}(\mathbf{x}_i \backslash \mathbf{r}_{i}^{(k)}), y_{i}^{*}))$ to become much larger than $\mathcal{L}_{\text{CE}}(\mathcal{F}_{\text{task}}(\mathbf{x}_i), y_{i}^{*})$, leading to arbitrarily negative losses.
To prevent this, we can use margin $m_c$ to impose a lower bound on $\mathcal{L}_{\text{comp-diff}}$, yielding the following \textit{margin criterion}: 

\vspace{-0.6cm}
\begin{equation}
\begin{split}
\mathcal{L}_{\text{comp-margin}} = \max(-m_c, \mathcal{L}_{\text{comp-diff}}) + m_c
\end{split}
\end{equation}
\vspace{-0.5cm}

Recall that suff measures the importance of tokens in $\mathbf{r}_{i}^{(k)}$ as how $p_{y_i}(\mathbf{x}_i)$ changes when they are the only tokens kept in $\mathbf{x}_i$.
Based on suff's definition, we want $p_{y_i}(\mathbf{r}_{i}^{(k)})$ to be higher than $p_{y_i}(\mathbf{x}_i)$, so lower suff is better.
For suff loss $\mathcal{L}_{\text{suff}}$, we define the difference and margin criteria analogously to $\mathcal{L}_{\text{comp}}$, using margin $m_s$ but the opposite sign for $\mathcal{L}_{\text{suff-diff}}$ (since lower suff is better):
\begin{equation}
\begin{split}
\mathcal{L}_{\text{suff-diff}} = \mathcal{L}_{\text{CE}}(\mathcal{F}_{\text{task}}(\mathbf{r}_{i}^{(k)}), y_{i}^{*})
 - \mathcal{L}_{\text{CE}}(\mathcal{F}_{\text{task}}(\mathbf{x}_i), y_{i}^{*})
\end{split}
\end{equation}
\vspace{-1.0cm}

\vspace{-0.3cm}
\begin{equation}
\begin{split}
\mathcal{L}_{\text{suff-margin}} = \max(-m_s, \mathcal{L}_{\text{suff-diff}}) + m_s
\end{split}
\end{equation}
\vspace{-0.6cm}


In our experiments, we find that the margin-based comp and suff criteria are effective (\textsection \ref{sec:experiments:ablation}), though others (\eg KL divergence, MAE) can be used too (\textsection \ref{sec:appendix:expl_objectives:faithfulness}).
Note that $\mathbf{r}_{i}^{(k)}$ is computed via top-$k\%$ thresholding (\textsection \ref{sec:background:rationale_extr}), so we also need to specify a set $K$ of threshold values.
We separately compute the comp and suff losses for each $k \in K$, then obtain the final comp and suff losses by averaging over all $k$ values via area-over-precision-curve (AOPC) \citep{deyoung2019eraser}.
To reflect this, we denote the comp and suff losses as $\mathcal{L}_{\text{comp},K}$ and $\mathcal{L}_{\text{suff},K}$, respectively.
Let $\alpha_f \mathcal{L}_{\text{faith}} = \alpha_c \mathcal{L}_{\text{comp},K} + \alpha_s \mathcal{L}_{\text{suff},K}$, where $\alpha_c$ and $\alpha_s$ are loss weights.
In this case, we can abstractly consider $\alpha_f$ as an aggregate loss weight for the faithfulness objectives.

\subsubsection{Plausibility}
\label{sec:method:expl_objectives:plausibility}
Plausibility is defined as how convincing a rationale (output by $\mathcal{F}_{\text{ext}}$) is to humans as an explanation for $\mathcal{F}_{\text{task}}$'s prediction on a given instance \citep{jacovi2020towards}.
Since $\mathcal{F}_{\text{task}}$'s predictions may change throughout its training, optimizing for plausibility should ideally involve continual human-in-the-loop feedback.
However, obtaining such human-in-the-loop feedback is prohibitive, so many works consider human-annotated gold rationales as a cheaper form of plausibility supervision \citep{deyoung2019eraser,narang2020wt5,jain2020learning}.
Even so, gold rationales $\mathbf{r}_{i}^{*}$ are generally only annotated with respect to the gold task label $y_{i}^{*}$ (as opposed to $\mathcal{F}_{\text{task}}$'s predicted label $y_{i}^{*}$, which cannot be known \textit{a priori}).
Consequently, if $y_{i}^{*} \neq y_{i}^{*}$, then gold rationale supervision may be noisy.

Still, this is not a significant issue if $\mathcal{F}_{\text{task}}$ is jointly trained with $\mathcal{F}_{\text{ext}}$.
In \method, $\mathcal{F}_{\text{task}}$ and $\mathcal{F}_{\text{ext}}$ are jointly trained to predict $y_{i}^{*}$ (via $\mathcal{L}_{\text{task}}$) and $\mathbf{r}_{i}^{*}$ (via $\mathcal{L}_{\text{plaus}}$), respectively, while $\mathcal{F}_{\text{task}}$ is also regularized (via $\mathcal{L}_{\text{faith}}$) such that its output $y_{i}^{*}$ aligns with $\mathcal{F}_{\text{ext}}$'s output $\mathbf{r}_{i}$.
In other words, $y_{i}^{*}$ may be an acceptable approximation of $y_{i}^{*}$ when training $\mathcal{F}_{\text{ext}}$ to predict $\mathbf{r}_{i}^{*}$ (which is based on $y_{i}^{*}$) because: (A) $\mathcal{F}_{\text{task}}$ is jointly trained such that its output $y_{i}^{*}$ approximates $y_{i}^{*}$, and (B) $\mathcal{F}_{\text{ext}}$ is also trained such that its output $\mathbf{r}_{i}$ aligns with $y_{i}^{*}$.
As a result, if gold rationale supervision is available, then we can optimize for plausibility via \method.
Specifically, given gold rationale $\mathbf{r}_{i}^{*}$ for input $\mathbf{x}_i$, plausibility optimization entails training $\mathcal{F}_{\text{ext}}$ to predict binary importance label $\mathbf{r}_{i}^{*,t}$ for each token $x_i^t$.
This is essentially binary token classification, so one natural choice for $\mathcal{L}_{\text{plaus}}$ is the token-level binary cross-entropy (BCE) criterion:

\vspace{-0.5cm}
\begin{equation} \label{eq:3}
\begin{split}
\mathcal{L}_{\text{plaus-BCE}} = - \sum_t \mathbf{r}_{i}^{*,t}\log(\mathcal{F}_{\text{ext}}(x_i^t))
\end{split}
\end{equation}
\vspace{-0.5cm}

Besides BCE loss, we can also consider other criteria like sequence-level KL divergence and linear loss.
See \textsection \ref{sec:appendix:expl_objectives:plausibility} for discussion of these and other plausibility criteria.

\subsection{Training and Inference}
\label{sec:method:training}



After setting $\mathcal{F}_{\text{ext}}$, $\mathcal{L}_{\text{faith}}$, and $\mathcal{L}_{\text{plaus}}$, we can move on to training $\mathcal{F}_{\text{task}}$ and $\mathcal{F}_{\text{ext}}$.
Since top-$k\%$ rationale binarization (\textsection \ref{sec:method:expl_objectives}) is not differentiable, by default, we cannot backpropagate $\mathcal{L}_{\text{faith}}$ through all of $\mathcal{F}_{\text{ext}}$'s parameters.
Thus, $\mathcal{F}_{\text{task}}$ is trained via $\mathcal{L}_{\text{task}}$ and $\mathcal{L}_{\text{faith}}$, while $\mathcal{F}_{\text{ext}}$ is only trained via $\mathcal{L}_{\text{plaus}}$.
This means $\mathcal{F}_{\text{ext}}$'s rationales $\mathbf{r}_i$ are indirectly optimized for faithfulness by regularizing $\mathcal{F}_{\text{task}}$ such that its behavior aligns with $\mathbf{r}_i$.
The exception is if we are using the SLM variant, where encoder $f_{\text{enc}}$ is shared by $\mathcal{F}_{\text{task}}$ and $\mathcal{F}_{\text{ext}}$.
In this case, $f_{\text{enc}}$ is optimized with respect to all losses, task head $f_{\text{task}}$ is optimized with respect to $\mathcal{L}_{\text{task}}$ and $\mathcal{L}_{\text{faith}}$, and extractor head $f_{\text{ext}}$ is optimized with respect to $\mathcal{L}_{\text{plaus}}$.
SLM is a simple way to approximate end-to-end training of $\mathcal{F}_{\text{task}}$ and $\mathcal{F}_{\text{ext}}$.
In contrast, past SPPs have used more complex methods like reinforcement learning \citep{lei2016rationalizing} and the reparameterization trick \citep{bastings2019interpretable}, whose training instability has been shown hurt task performance \citep{jain2020learning}.

Now, we summarize the full learning objective.
Given that cross-entropy loss $\mathcal{L}_{\text{task}} = \mathcal{L}_{\text{CE}}(\mathcal{F}_{\text{task}}(\mathbf{x}_i), y_{i}^{*})$ is used to train $\mathcal{F}_{\text{task}}$ to predict $y_{i}^{*}$, the full learning objective is: 
\begin{equation}
\begin{split}
\mathcal{L} = \hspace{1mm} &\mathcal{L}_{\text{task}} + \alpha_{f} \mathcal{L}_{\text{faith}} + \alpha_{p} \mathcal{L}_{\text{plaus}} \\
= \hspace{1mm} &\mathcal{L}_{\text{task}} + \alpha_{c} \mathcal{L}_{\text{comp},K} + \alpha_{s} \mathcal{L}_{\text{suff},K} + \alpha_{p} \mathcal{L}_{\text{plaus}}.
\end{split}
\end{equation}
During inference, we use $\mathcal{F}_{\text{task}}$ to predict $y_{i}^{*}$, then use $\mathcal{F}_{\text{ext}}$ to output $\mathbf{r}_i$ for $\mathcal{F}_{\text{task}}$'s predicted label $y_i$.



\section{Experiments} 
\label{sec:experiments}

We present empirical results showing \method's effectiveness in trading off faithfulness, plausibility, and task performance during rationale extractor optimization.
First, our main experiments compare rationale extraction methods w.r.t. all three desiderata (\textsection \ref{sec:experiments:main}).
Second, we perform various ablation studies to verify our design choices for \methodsp  (\textsection \ref{sec:experiments:ablation}).
Third, we present experiments showing \method's strong data efficiency, w.r.t. limited gold rationale supervision  (\textsection \ref{sec:experiments:gold_eff}) and zero-shot faithfulness transfer  (\textsection \ref{sec:experiments:zs}).
Fourth, to account for the limitations of gold-rationale-based plausibility evaluation, we conduct a user study to further demonstrate the improved plausibility of \method-extracted rationales  (\textsection \ref{sec:experiments:user_study}).

\vspace{-0.1cm}
\subsection{Evaluation Protocol}
\label{sec:experiments:eval}

\subsubsection{Datasets}
\label{sec:experiments:eval:datasets}
We primarily experiment with the SST (sentiment analysis) \citep{socher2013recursive, carton2020evaluating}, Movies (sentiment analysis) \citep{zaidan2008modeling}, CoS-E (commonsense question answering) \citep{rajani2019explain}, MultiRC (reading comprehension) \citep{khashabi2018looking}, and e-SNLI (natural language inference) \citep{camburu2018snli} datasets, all of which have gold rationale annotations.
The rationale-annotated version of SST was obtained from \citet{carton2020evaluating}, while the latter four datasets were obtained from the ERASER benchmark \citep{deyoung2019eraser}.
For the zero-shot faithfulness transfer experiments (\textsection \ref{sec:experiments:zs}), we consider five additional datasets, which are described further in \textsection \ref{sec:experiments:zs}.
For more details, please refer to \textsection \ref{sec:appendix:datasets}.

\subsubsection{Metrics}
\label{sec:experiments:eval:metrics}
To measure faithfulness, plausibility, and task performance, we use the metrics from the ERASER benchmark \citep{deyoung2019eraser}.
For faithfulness, we use comp and suff, for $k = [1, 5, 10, 20, 50]$ \citep{deyoung2019eraser}. 
For plausibility, we use area under precision-recall curve (AUPRC) and token F1 (TF1) to measure similarity to gold rationales \citep{deyoung2019eraser, narang2020wt5}.
For task performance, we follow the dataset-specific metrics used in ERASER: accuracy for SST and CoS-E; macro F1 for Movies, MultiRC, and e-SNLI \citep{deyoung2019eraser}.
That is, we only use one task performance metric per dataset.

\paragraph{Normalized Relative Gain (NRG)}
After computing these raw metrics for faithfulness, plausibility, and task performance, we would like to compare different rationale extraction methods w.r.t. all three desiderata.
However, aggregating the raw metrics across the three desiderata may not be straightforward.
In light of this, we introduce the Normalized Relative Gain (NRG) metric, which is based on the Average Relative Gain (ARG) metric \citep{ye2021crossfit} and min-max scaling.
For each raw metric, NRG transforms all raw scores to normalized scores in $[0, 1]$ (higher is better).
After all raw metrics are in the same $[0, 1]$ space, we can simply aggregate them via averaging.

Concretely, for each raw metric (\eg comp, suff, AUPRC, accuracy), we are given a set of raw scores $Z = \{z_1, z_2, ...\}$.
Each raw score $z_i \in Z$ corresponds to a different rationale extraction method $i$. 
$\text{NRG}(z_i)$ captures $z_i$'s relative gain over the worst score in $Z$, normalized w.r.t. score range $\max(Z) - \min(Z)$.
The definition of ``worst score" depends on whether higher or lower raw scores are better for the given metric.
If higher raw scores are better (\eg comp, AUPRC, accuracy), then the worst score would be $\min(Z)$, which yields: $\text{NRG}(z_i) = \frac{z_i - \min(Z)}{\max(Z) - \min(Z)}$.
If lower values are better (\eg sufficiency), then the worst score would be $\max(Z)$, which yields: $\text{NRG}(z_i) = \frac{\max(Z)-z_i}{\max(Z)-\min(Z)}$.

After computing the individual NRG for each raw metric, we obtain the desiderata NRG scores by averaging the individual NRG scores within each desideratum.
Let FNRG, PNRG, and TNRG be the desiderata NRG scores for faithfulness, plausibility, and task performance, respectively.
FNRG is the average of the individual NRG scores for comp and suff; PNRG is the average of the individual NRG scores for AUPRC and TF1; and TNRG is just the individual NRG for the task performance metric (since there is only one task performance metric per dataset).
Finally, to summarize all of the raw metrics as a single score, we compute the composite NRG (CNRG) by averaging the three desiderata NRG scores: $\text{CNRG} = \frac{\text{FNRG} + \text{PNRG} + \text{TNRG}}{3}$.
By default, we compute CNRG as an unweighted average of the three desiderata NRG scores, under the assumption that all three desiderata are equally important.
On the other hand, for situations where certain desiderata are more important than others, we can also compute CNRG as a weighted average.

Generally, the computation of NRG should involve globally aggregating the raw metrics across all available methods, which is done in the main results (\textsection \ref{sec:experiments:main}).
However, for a number of more focused experiments (\textsection \ref{sec:experiments:ablation} and \ref{sec:experiments:zs}), only a subset of the available methods are considered.
Thus, for these experiments, we report the raw metrics instead of NRG (Tables \ref{tab:ablations} and \ref{tab:transfer}).


\subsubsection{Results Reporting}
For all results, we report the average over three seeds and five $k$ faithfulness thresholds (\ie $k = [1, 5, 10, 20, 50]$), a total of 15 settings.
We denote each \methodsp configuration with a parenthetical ``([\textit{rationale extractor}]-[\textit{explainability objectives}])''.
For the rationale extractor, AA, DLM, and SLM denote attribution algorithm, Dual LM, and Shared LM, respectively.
For explainability objectives, F, P, and FP denote faithfulness, plausibility, and faithfulness+plausibility, respectively.
For example, DLM-FP means Dual LM with faithfulness+plausibility objectives.

\subsection{Baselines}
We consider a wide range of representative rationale extraction baselines, spanning three key categories.
Note that some methods do not assume access to gold rationales, which prevents such methods from optimizing for plausibility.
This means that not all of the methods are directly comparable.
Therefore, when comparing methods, we generally group them by whether they use optimize for plausibility and only compare methods within the same group.

The first category is \textbf{vanilla attribution algorithm (AA)}, which does not involve training $\mathcal{F}_{\text{ext}}$ and is applied post hoc (\ie they do not impact $\mathcal{F}_{\text{task}}$'s training). 
Included baselines from this category are: Gradient (Grad) \citep{simonyan2013deep}, Input*Gradient (Input*Grad) \citep{denil2014extraction}, DeepLIFT \citep{lundberg2017unified}, and Integrated Gradients (IG) \citep{sundararajan2017axiomatic}.
These four baselines are among the most popular AAs in the explainability literature \citep{luo2021local, pruthi2020evaluating}.
In the results, we denote these baselines as ``AA ([\textit{AA name}])'', e.g., AA (IG).

The second category is \textbf{AA-based training}, which uses AAs in some way to train a learned $\mathcal{F}_{\text{ext}}$.
One baseline in this category is L2E \citep{situ2021learning}, which distills knowledge from an AA to an LM-based $\mathcal{F}_{\text{ext}}$.
Specifically, after training $\mathcal{F}_{\text{task}}$, then using an AA to extract rationales for $\mathcal{F}_{\text{task}}$, L2E entails training $\mathcal{F}_{\text{ext}}$ to output rationales that are similar to the AA's rationales.
Another baseline is SGT \citep{ismail2021improving}, which uses a suff-based criterion to regularize $\mathcal{F}_{\text{task}}$, such that the AA yields faithful rationales for $\mathcal{F}_{\text{task}}$.
We also consider a variant called SGT+P, which augments SGT with plausibility optimization via gold rationales.
For all baselines in this category, we use IG as the AA.

The third category is \textbf{select-predict pipeline (SPP)}, where $\mathcal{F}_{\text{task}}$ (predictor) only takes input tokens chosen via $\mathcal{F}_{\text{ext}}$'s (selector) rationale output.
One baseline in this category is FRESH \citep{jain2020learning}, which trains $\mathcal{F}_{\text{task}}$ and $\mathcal{F}_{\text{ext}}$ separately.
For FRESH, we use a stronger variant (compared to those in the FRESH paper) where IG rationales are directly provided to the predictor, rather than output by a trained $\mathcal{F}_{\text{ext}}$.
Another baseline is A2R \citep{yu2021understanding}, a recently proposed SPP which aims to improve $\mathcal{F}_{\text{task}}$'s task performance by regularizing $\mathcal{F}_{\text{task}}$ with an attention-based predictor that uses the full input.
Also, we introduce FRESH+P and A2R+P, which respectively augment FRESH and A2R with plausibility optimization.

\subsection{Implementation Details}
\label{sec:experiments:implementation}
For the LM architecture of $\mathcal{F}_{\text{task}}$ and $\mathcal{F}_{\text{ext}}$, we use BigBird-Base \citep{zaheer2020big} in all of our experiments, in order to handle input sequences of up to 4096 tokens.
For all AA-based methods besides vanilla AA, we use the IG \citep{sundararajan2017axiomatic} AA, which has been commonly adopted in the explainability literature \citep{pruthi2020evaluating, sanyal2021discretized, ismail2021improving}.
By default, IG involves 50 steps (\ie backward passes) per instance \citep{kokhlikyan2020captum}.
However, 50-step IG is prohibitive when regularizing the LM via IG, so we instead use 3-step IG in both training and evaluation.
We empirically justify our usage of 3-step IG in \textsection \ref{sec:appendix:compute}.

For all experiments, we use a learning rate of $2\mathrm{e}{-5}$ and effective batch size of 32.
We train for a maximum of 10 epochs, with early stopping patience of 5 epochs.
For the gold rationale efficiency experiments (\textsection \ref{sec:experiments:gold_eff}, \textsection \ref{sec:appendix:gold_eff}), we use a batching factor $\beta$ (\textsection \ref{sec:appendix:gold}) of 2.
We only tune faithfulness and plausibility loss weights, sweeping $\alpha_c = [0.5, 0.7, 1.0]$, $\alpha_s = [0.5, 0.7, 1.0]$, and $\alpha_p = [0.5, 0.7, 1.0]$.
We find that $\alpha_c = 0.5$ and $\alpha_s = 0.5$ typically yield the best performance.
For each method variant, we tuned hyperparameters w.r.t. dev CNRG, computed across all hyperparameter configurations for the variant.
All of our experiments are implemented using PyTorch \citep{paszke2019pytorch}, Lightning \citep{Falcon_PyTorch_Lightning_2019}, and Hugging Face Transformers \citep{wolf2019huggingface}.






\begin{figure*}[h!]
	\centering
	\includegraphics[width=\textwidth]{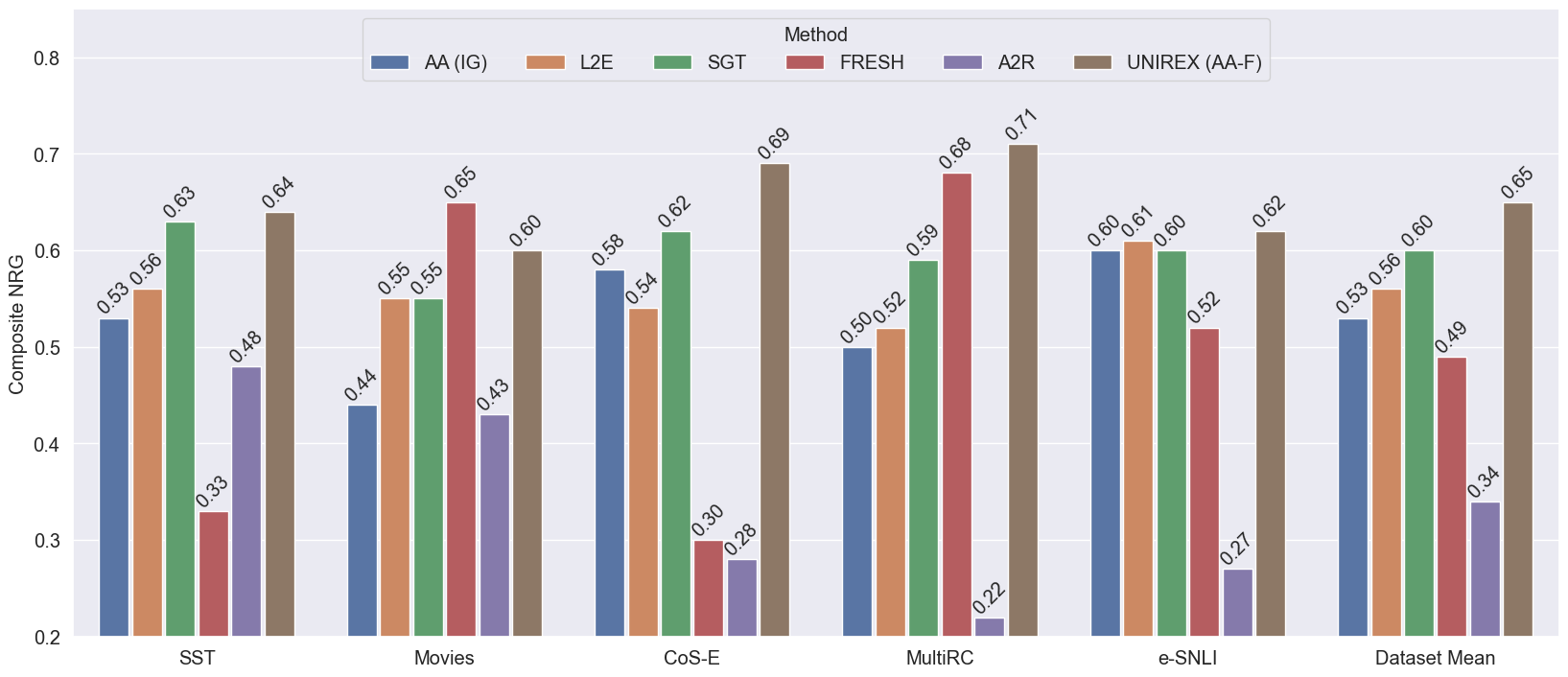}
	\caption{\small \textbf{Composite NRG Comparison (without Plausibility Optimization).}
		The composite NRG (CNRG) is the mean of the three desiderata NRG scores.
		For each dataset, we use CNRG to compare rationale extraction methods that \textit{do not} optimize for plausibility.
		Overall, \methodsp (AA-F) achieves the best CNRG on Dataset Mean (and on all datasets except Movies), showing the effectiveness of \method's faithfulness optimization.
		On Dataset Mean, \methodsp (AA-F) beats the strongest baseline (\ie SGT) by 9.2\%.
	}
	\label{fig:main:composite_1}
\end{figure*}

\begin{figure*}[h!]
	\centering
	\includegraphics[width=\textwidth]{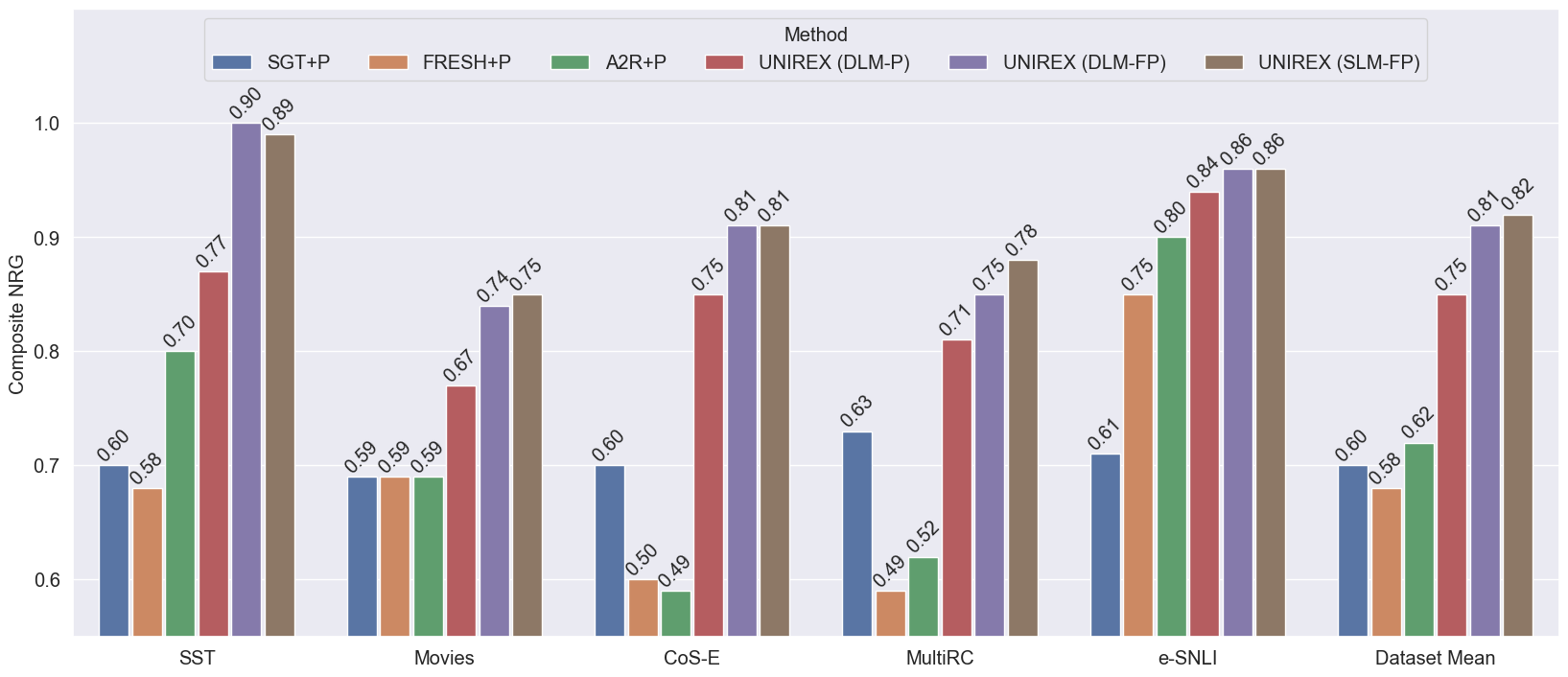}
	\caption{\small \textbf{Composite NRG Comparison (with Plausibility Optimization).}
		The composite NRG (CNRG) is the mean of the three desiderata NRG scores.
		For each dataset, we use CNRG to compare rationale extraction methods that \textit{do} optimize for plausibility.
		Overall, \methodsp (SLM-FP) and \methodsp (DLM-FP) achieve the best CNRG -- both beating the strongest baseline (\ie A2R+P) by over 30\% on Dataset Mean -- demonstrating \method's ability to jointly optimize $\mathcal{F}_{\text{task}}$ and $\mathcal{F}_{\text{ext}}$ for all three desiderata.
	}
	\label{fig:main:composite_2}
\end{figure*}

\begin{figure*}[h!]
	\centering
	\includegraphics[width=\textwidth]{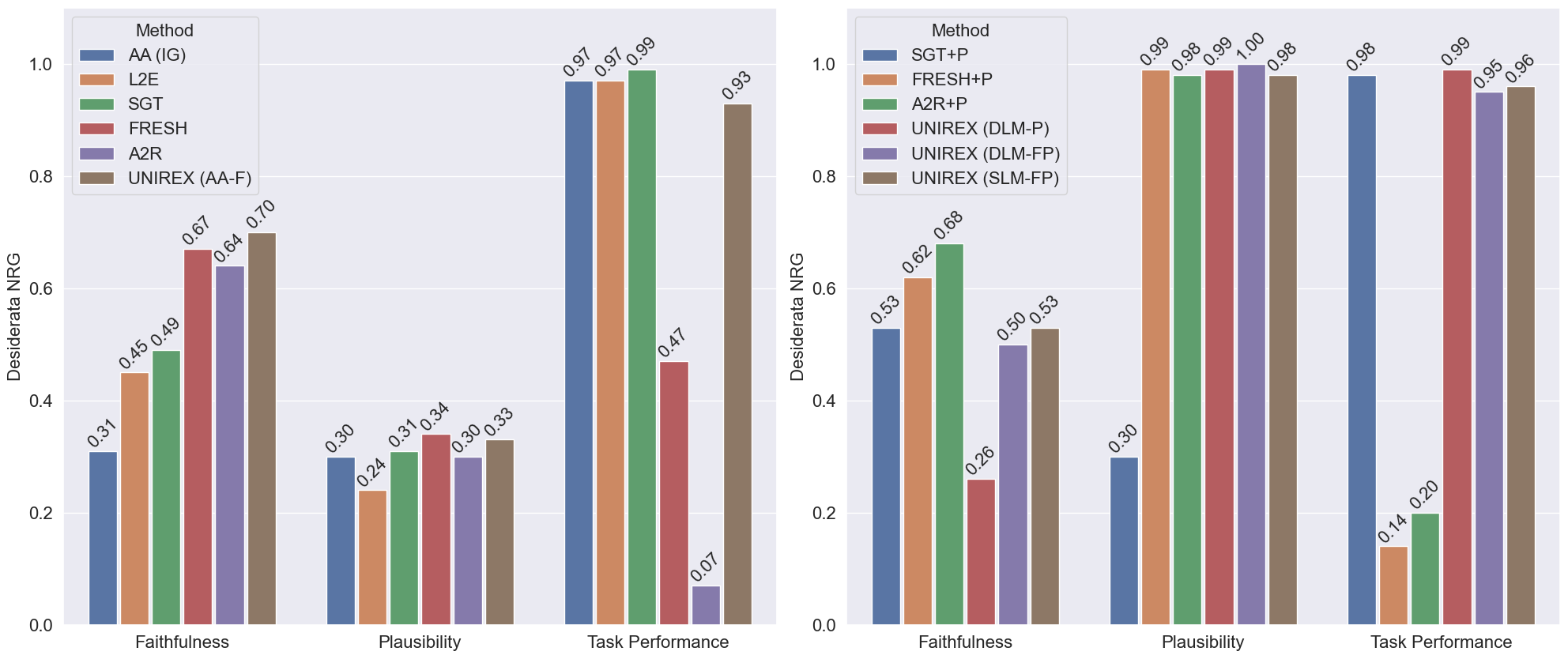}
	\caption{\small \textbf{Desiderata NRG Comparison.} 
		For each rationale extraction method, we show the desiderata NRG for faithfulness (FNRG), plausibility (PNRG), and task performance (TNRG), averaged over all datasets.
		\textbf{Left}: This plot compares methods \textit{without} plausibility optimization.
		\methodsp (AA-F)'s FNRG is highest, while its TNRG is close to highest.
		Meanwhile, baselines with high FNRG (\ie FRESH, A2R) have low TNRG, while baselines with high TNRG (\ie AA (IG), L2E, SGT) have low FNRG.
		\textbf{Right}: This plot compares methods \textit{with} plausibility optimization.
		\methodsp (DLM-FP) and \methodsp (SLM-FP) have moderate FNRG, but the highest (or near-highest) PNRG and TNRG.
		Meanwhile, baselines with high FNRG (\ie FRESH+P, A2R+P) have low TNRG, while baselines with high TNRG (\ie SGT+P) have low PNRG.
	}
	\label{fig:main:desiderata}
\end{figure*}

\subsection{Main Results}
\label{sec:experiments:main}
Figs. \ref{fig:main:composite_1}-\ref{fig:main:desiderata} display the main results, in terms of NRG.
For conciseness, we omit AA (Grad), AA (Input*Grad), and AA (DeepLIFT) results from these NRG figures, since AA (IG) is representative of vanilla AA methods.
Please refer to \textsection \ref{sec:appendix:raw_results} for all raw and NRG empirical results.

In Figs. \ref{fig:main:composite_1}-\ref{fig:main:composite_2}, we use CNRG to compare rationale extraction methods for each dataset.
Here, the Dataset Mean group reports the mean CNRG across all datasets.
First, Fig. \ref{fig:main:composite_1} compares methods that \textit{do not} optimize for plausibility (since they do not have access to gold rationales).
Overall, we find that \methodsp (AA-F) achieves the best CNRG on Dataset Mean (and on all datasets except Movies), showing the effectiveness of \method's faithfulness optimization.
On Dataset Mean, \methodsp (AA-F) beats the strongest baseline (\ie SGT) by 9.2\%.
Second, Fig. \ref{fig:main:composite_2} compares methods that \textit{do} optimize for plausibility.
Overall, we find that \methodsp (SLM-FP) and \methodsp (DLM-FP) achieve the best CNRG -- both beating the strongest baseline (\ie A2R+P) by over 30\% on Dataset Mean -- demonstrating \method's ability to jointly optimize $\mathcal{F}_{\text{task}}$ and $\mathcal{F}_{\text{ext}}$ for all three desiderata.
Meanwhile, \methodsp (DLM-P) performs slightly worse but still significantly better than all baselines, showing the effectiveness of \method's plausibility optimization.


\begin{table*}[ht!]
	\centering
	\scalebox{0.8}{
		\begin{tabular}{cccccc}
			\toprule
			\multirow{2}{*}{\textbf{Ablation}} & \multirow{2}{*}{\textbf{\methodsp Config}} & \multicolumn{2}{c}{\textbf{Faithfulness}} & \textbf{Plausibility} & \textbf{Performance}\\
			\cmidrule(lr){3-4} \cmidrule(lr){5-5} \cmidrule(lr){6-6}
			& & Comp ($\uparrow$) & Suff ($\downarrow$) & AUPRC ($\uparrow$) & Acc ($\uparrow$) \\
			\midrule
			\multirow{4}{*}{Ext Type (F)} & {AA-F (Rand)} & 0.171~($\pm$0.040) & 0.327~($\pm$0.050) & 44.92~($\pm$0.00) & \textbf{94.05}~($\pm$0.35)  \\
			
			& {AA-F (Gold)} & 0.232~($\pm$0.088) & 0.249~($\pm$0.021) & \textbf{100.00}~($\pm$0.00) & 93.81~($\pm$0.54) \\
			
			& {AA-F (Inv)} & 0.242~($\pm$0.010) & 0.357~($\pm$0.019) & 20.49~($\pm$0.00) & 93.47~($\pm$1.81)  \\
			
			& {AA-F (IG)} & \textbf{0.292}~($\pm$0.051) & \textbf{0.171}~($\pm$0.038) & 48.13~($\pm$1.14) & 92.97~($\pm$0.44) \\
			
			\midrule
			
			\multirow{4}{*}{Ext Type (FP)} & {AA-FP (Sum)} & 0.296~($\pm$0.067) & 0.185~($\pm$0.048) & 47.60~($\pm$2.44) & 93.25~($\pm$0.45) \\
			
			& {AA-FP (MLP)} & 0.285~($\pm$0.051) & 0.197~($\pm$0.100) & 54.82~($\pm$1.97) & 93.23~($\pm$0.92) \\
			
			& {DLM-FP} & \textbf{0.319}~($\pm$0.090) & 0.167~($\pm$0.036) & \textbf{85.80}~($\pm$0.74) & \textbf{93.81}~($\pm$0.18) \\
			
			& {SLM-FP} & 0.302~($\pm$0.039) & \textbf{0.113}~($\pm$0.013) & 82.55~($\pm$0.84) & 93.68~($\pm$0.67) \\
			
			\midrule
			
			\multirow{3}{*}{Comp/Suff Loss} & {SLM-FP (Comp)} & \textbf{0.350}~($\pm$0.048) & 0.310~($\pm$0.049) & 82.79~($\pm$0.62) & 93.59~($\pm$0.11) \\
			
			& {SLM-FP (Suff)} & 0.166~($\pm$0.003) & 0.152~($\pm$0.012) & \textbf{83.74}~($\pm$0.84) & \textbf{94.16}~($\pm$0.39) \\
			
			& {SLM-FP (Comp+Suff)} & 0.302~($\pm$0.039) & \textbf{0.113}~($\pm$0.013) & 82.55~($\pm$0.84) & 93.68~($\pm$0.67) \\
			
			\midrule
			
			\multirow{3}{*}{Suff Criterion} & {SLM-FP (KL Div)} & \textbf{0.306}~($\pm$0.098) & 0.131~($\pm$0.005) & 82.62~($\pm$0.88) & 93.06~($\pm$0.25) \\
			
			& {SLM-FP (MAE)} & 0.278~($\pm$0.058) & 0.143~($\pm$0.008) & \textbf{82.66}~($\pm$0.61) & \textbf{93.78}~($\pm$0.13) \\
			
			& {SLM-FP (Margin)}& 0.302~($\pm$0.039) & \textbf{0.113}~($\pm$0.013) & 82.55~($\pm$0.84) & 93.68~($\pm$0.67) \\
			
			\midrule
			
			\multirow{3}{*}{SLM Ext Head} & {SLM-FP (Linear)} & 0.302~($\pm$0.039) & \textbf{0.113}~($\pm$0.013) & 82.55~($\pm$0.84) & \textbf{93.68}~($\pm$0.67) \\
			
			& {SLM-FP (MLP-2048-2)} & \textbf{0.323}~($\pm$0.071) & 0.144~($\pm$0.012) & 83.82~($\pm$0.77) & 93.67~($\pm$0.18) \\
			
			& {SLM-FP (MLP-4096-3)} & 0.295~($\pm$0.057) & 0.154~($\pm$0.027) & \textbf{84.53}~($\pm$0.61) & 93.19~($\pm$0.79) \\
			
			\bottomrule 
		\end{tabular}
	}
	\caption{\small \textbf{\methodsp Ablation Studies on SST}
	}
	\label{tab:ablations}
	\vspace{-0.2cm}
\end{table*}

Fig. \ref{fig:main:desiderata} compares rationale extraction methods w.r.t. the desiderata NRG, averaged over all datasets.
First, Fig. \ref{fig:main:desiderata} (\textbf{left}) compares rationale extraction methods \textit{without} plausibility optimization, so PNRG is low for all methods here.
Here, we see that \methodsp (AA-F)'s FNRG is highest, while its TNRG is close to highest.
As shown in Fig. \ref{fig:main:composite_1}, \methodsp (AA-F) achieves the best composite NRG (CNRG) because \methodsp training enables effective balancing of faithfulness and task performance.
On the other hand, baselines with high FNRG (\ie FRESH, A2R) have low TNRG, while baselines with high TNRG (\ie AA (IG), L2E, SGT) have low FNRG.
Second, Fig. \ref{fig:main:desiderata} (\textbf{right}) compares rationale extraction methods \textit{with} plausibility optimization.
Here, we see that \methodsp (DLM-FP) and \methodsp (SLM-FP) have moderate FNRG, but the highest (or near-highest) PNRG and TNRG.
Meanwhile, \methodsp (DLM-P) achieves the highest PNRG and TNRG, but the worst FNRG, since \methodsp (DLM-P) does not optimize for faithfulness.
As shown in Fig. \ref{fig:main:composite_2}, \methodsp (DLM-FP) and \methodsp (SLM-FP) achieve the best CNRG because \methodsp training enables effective balancing of faithfulness and task performance.
Meanwhile, baselines with high FNRG (\ie FRESH+P, A2R+P) have low TNRG, while baselines with high TNRG (\ie SGT+P) have low PNRG.

\subsection{Ablation Studies}
\label{sec:experiments:ablation}

We present five ablation studies to validate the effectiveness of our \methodsp design choices.
The results of these ablation studies are displayed in Table \ref{tab:ablations}, where each of the five sections contains results for a different ablation.
Thus, all numbers within the same section (ablation) and column (metric) are comparable.

\paragraph{Extractor Type (F)}
In the Ext Type (F) section, we compare four heuristic rationale extractors, using AA-F.
In this case, besides task performance, we can only optimize (the task model) for faithfulness.
Rand uses random importance scores, Gold directly uses the gold rationales, Inv uses the inverse of the gold rationales, and IG uses IG rationales.
All heuristics yield similar task performance, but IG dominates on all faithfulness metrics.
This makes sense because IG is computed using $\mathcal{F}_{\text{task}}$'s inputs/parameters/outputs, while the others do not have this information.
For plausibility, Gold is the best, Inv is the worst, and Rand and IG are about the same, as none of the heuristics are optimized for plausibility.

\paragraph{Extractor Type (P)}
In the Ext Type (FP) section, we compare four learned rationale extractors.
In this case, besides task performance, we can optimize for both faithfulness and plausibility.
By default, attribution algorithms' dimension scores are pooled into token scores via sum pooling.
AA-FP (Sum) uses IG with sum pooling, while AA-FP (MLP) replaces the sum pooler with a MLP-based pooler to increase capacity for plausibility optimization.
Task performance for all four methods is similar, AA-FP (Sum) dominates on faithfulness, and DLM-FP and SLM-FP dominate on plausibility.
AA-FP (MLP) does not perform as well on faithfulness but slightly improves on plausibility compared to AA-FP (Sum).

\paragraph{Comp/Suff Losses}
The Comp/Suff Loss section compares different combinations of Comp and Suff losses, using SLM-FP.
Note that SLM-FP (Comp+Suff) is equivalent to SLM-FP shown in other tables/sections.
As expected, SLM-FP (Comp) does best on Comp, but SLM-FP (Comp+Suff) actually does best on Suff.
Meanwhile, SLM-FP, (Suff) does second-best on Suff but is much worse on Comp.
This shows that Comp and Suff are complementary for optimization.

\paragraph{Suff Criterion}
The Suff Criterion section compares different Suff criteria, using SLM-FP.
SLM-FP (KLDiv) uses the KL divergence criterion, SLM-FP (MAE) uses the MAE criterion, and SLM-FP (Margin) uses the margin criterion.
SLM-FP (Margin) is equivalent to SLM-FP in other tables/sections.
All criteria yield similar performance and plausibility, while Margin is slightly better on faithfulness.

\paragraph{SLM Extractor Head}
The SLM Ext Head section compares different extractor heads, using SLM-FP.
Linear is the default choice and uses a linear layer. 
MLP-2048-2 uses a MLP with two 2048-dim hidden layers.
MLP-4096-3 uses a MLP with three 4096-dim hidden layers.
All three output head types yield similar performance, but decreasing head capacity yields better faithfulness, while increasing head capacity heads yields better plausibility.
This trades off faithfulness and plausibility, although larger heads will be more compute-intensive.

\begin{figure}[t!]
	\centering
	\includegraphics[width=0.48\textwidth]{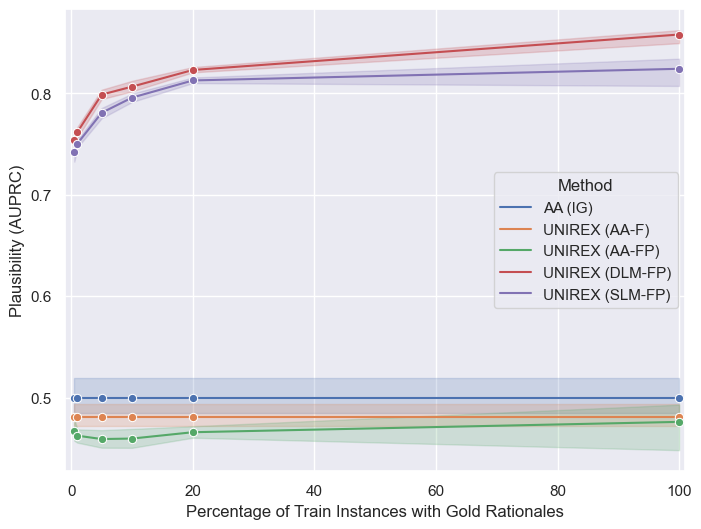}
	\caption{\small \textbf{Gold Rationale Efficiency on SST.} As shown in this plot, \methodsp (DLM-FP) and \methodsp (SLM-FP) are able to achieve high plausibility performance, even with a very small percentage of training instances with gold rationale annotations.}
	\vspace{-0.2cm}
	\label{fig:gold_eff:sst}
\end{figure}
\vspace{-0.3cm}

\subsection{Gold Rationale Efficiency}
\label{sec:experiments:gold_eff}

\methodsp supports arbitrary amounts of gold rationale supervision, allowing plausibility optimization even in low-resource settings.
In Fig. \ref{fig:gold_eff:sst}, we compare plausibility (w.r.t. AUPRC) for $\gamma = [0.5, 1, 5, 10, 20, 100]$ (\ie $\%$ of train instances with gold rationales).
We compare AA (IG) and four \methodsp variants (AA-F, AA-FP, DLM-FP, SLM-FP), with standard deviation shown by the error bands.
First, AA (IG) and \methodsp (AA-F) do not optimize for plausibility via gold rationales, so their low AUPRC scores are constant for all $\gamma$.
Second, \methodsp (AA-FP)'s AUPRC varies directly with $\gamma$ at a modest rate, but is still always lower than AA (IG)'s and \methodsp (AA-F)'s AUPRC.
\methodsp (AA-FP)'s plausibility optimization is not effective, since plausibility optimization (\ie learning to generate human-like rationales) typically requires high learning capacity, yet AAs do not have any learnable parameters.  
Third, \methodsp (DLM-FP) and \methodsp (SLM-FP) dominate across all $\gamma$ values, with AUPRC slowly decreasing as $\gamma$ decreases.
Even at $\gamma=0.5$, they can still achieve high AUPRC scores of around 0.75.
This suggests that \methodsp's gold rationale batching procedure (\textsection \ref{sec:appendix:gold}) is helpful for learning from minimal gold rationale supervision, thus enabling effective plausibility optimization.
In addition to these results on SST, see Fig. \ref{fig:gold_eff:cose} for similar results on CoS-E.



\begin{table*}[ht!]
	\centering
	\scalebox{0.8}{
		\begin{tabular}{ccccccc}
			\toprule
			\multirow{2}{*}{\textbf{Task}} & \multirow{2}{*}{\textbf{Dataset}} & \multirow{2}{*}{\textbf{Method}} & \multicolumn{3}{c}{\textbf{Faithfulness}} & \textbf{Task Performance} \\
			\cmidrule(lr){4-6} \cmidrule(lr){7-7}
			& & & CSD ($\uparrow$) & Comp ($\uparrow$) & Suff ($\downarrow$) & Perf ($\uparrow$) \\
			
			\midrule
			
			\multirow{19}{*}{SA} & \multirow{6}{*}{SST} & {AA (IG)} & -0.138~($\pm$0.040) & 0.119~($\pm$0.009) & 0.258~($\pm$0.031) & 93.81~($\pm$0.55) \\
			& & {AA-F (Rand)} & -0.156~($\pm$0.018) & 0.171~($\pm$0.040) & 0.327~($\pm$0.050) & 94.05~($\pm$0.35) \\
			& & {\methodsp (AA-F)} & 0.120~($\pm$0.055) & 0.292~($\pm$0.051) & 0.171~($\pm$0.038) & 92.97~($\pm$0.44) \\
			& & {\methodsp (DLM-P)} & -0.113~($\pm$0.040) & 0.142~($\pm$0.008) & 0.255~($\pm$0.007) & \textbf{94.86}~($\pm$0.41) \\
			& & {\methodsp (DLM-FP)} & 0.151~($\pm$0.056) & \textbf{0.319}~($\pm$0.090) & 0.167~($\pm$0.036) & 93.81~($\pm$0.54) \\
			& & {\methodsp (SLM-FP)} & \textbf{0.189}~($\pm$0.030) & 0.302~($\pm$0.039) & \textbf{0.113}~($\pm$0.013)& 93.68~($\pm$0.67)  \\
			
			\cmidrule{2-7}
			
			& \multirow{6}{*}{Yelp} & {AA (IG)} & -0.149~($\pm$0.028) & 0.069~($\pm$0.004) & 0.219~($\pm$0.028) & 92.50~($\pm$2.07) \\
			& & {AA-F (Rand)} & -0.177~($\pm$0.056) & 0.127~($\pm$0.022) & 0.305~($\pm$0.060) & 86.27~($\pm$7.88) \\
			& & {\methodsp (AA-F)} & 0.013~($\pm$0.036) & 0.138~($\pm$0.078) & 0.126~($\pm$0.059) & 83.93~($\pm$13.20) \\
			& & {\methodsp (DLM-P)} & -0.004~($\pm$0.028) & 0.138~($\pm$0.009) & 0.143~($\pm$0.023) & \textbf{93.33}~($\pm$0.90) \\
			& & {\methodsp (DLM-FP)} & \textbf{0.169}~($\pm$0.060) & \textbf{0.265}~($\pm$0.094) & 0.097~($\pm$0.033) & 92.37~($\pm$0.46) \\
			& & {\methodsp (SLM-FP)} & 0.114~($\pm$0.056) & 0.175~($\pm$0.055) & \textbf{0.060}~($\pm$0.001) & 86.60~($\pm$1.57) \\
			
			\cmidrule{2-7}
			
			& \multirow{6}{*}{Amazon} & {AA (IG)} & -0.148~($\pm$0.038) & 0.076~($\pm$0.010) & 0.224~($\pm$0.037) & \textbf{91.13}~($\pm$0.28) \\
			& & {AA-F (Rand)} & -0.166~($\pm$0.035) & 0.120~($\pm$0.021) & 0.286~($\pm$0.035) & 81.52~($\pm$6.95) \\
			& & {\methodsp (AA-F)} & 0.057~($\pm$0.106) & 0.130~($\pm$0.077) & 0.073~($\pm$0.039) & 77.90 ~($\pm$13.12) \\
			& & {\methodsp (DLM-P)} & -0.017~($\pm$0.027) & 0.142~($\pm$0.001) & 0.158~($\pm$0.026) & 90.92~($\pm$0.93) \\
			& & {\methodsp (DLM-FP)} & \textbf{0.133}~($\pm$0.039) & \textbf{0.232}~($\pm$0.072) & 0.098~($\pm$0.033) & 89.35~($\pm$2.22) \\
			& & {\methodsp (SLM-FP)} & 0.097~($\pm$0.027) & 0.147~($\pm$0.012) & \textbf{0.050}~($\pm$0.017) & 81.82~($\pm$7.62) \\
			
			\midrule
			
			\multirow{6}{*}{HSD} & \multirow{6}{*}{Stormfront} & {AA (IG)} & -0.109~($\pm$0.053) & 0.135~($\pm$0.010) & 0.245~($\pm$0.059) & 10.48~($\pm$1.66) \\
			& & {AA-F (Rand)} & -0.147~($\pm$0.021) & 0.150~($\pm$0.020) & 0.297~($\pm$0.005) & \textbf{10.66}~($\pm$2.86) \\
			& & {\methodsp (AA-F)} & \textbf{0.127}~($\pm$0.015) & \textbf{0.219}~($\pm$0.009) & 0.092~($\pm$0.025) & 10.36~($\pm$1.94) \\
			& & {\methodsp (DLM-P)} & -0.125~($\pm$0.092) & 0.122~($\pm$0.008) & 0.246~($\pm$0.099) & 10.10~($\pm$1.73) \\
			& & {\methodsp (DLM-FP)} & 0.052~($\pm$0.027) & 0.167~($\pm$0.084) & 0.115~($\pm$0.059) & 10.37~($\pm$2.66) \\
			& & {\methodsp (SLM-FP)} & 0.049~($\pm$0.041) & 0.110~($\pm$0.039) & \textbf{0.062}~($\pm$0.043) & 4.51~($\pm$1.87) \\
			
			\midrule
			
			\multirow{6}{*}{OSD} & \multirow{6}{*}{OffenseEval} & {AA (IG)} & -0.146~($\pm$0.044) & 0.097~($\pm$0.009) & 0.244~($\pm$0.052) & 33.51~($\pm$0.99)  \\
			& & {AA (Rand)} & -0.148~($\pm$0.046) & 0.101~($\pm$0.020) & 0.249~($\pm$0.065) & 34.08~($\pm$2.34) \\
			& & {\methodsp (AA-F)} & -0.029~($\pm$0.040) & 0.074~($\pm$0.040) & 0.102~($\pm$0.024) & 32.62~($\pm$4.85) \\
			& & {\methodsp (DLM-P)} & -0.102~($\pm$0.073) & 0.112~($\pm$0.010) & 0.214~($\pm$0.081) & 33.67~($\pm$1.01) \\
			& & {\methodsp (DLM-FP)} & \textbf{0.053}~($\pm$0.012) & \textbf{0.140}~($\pm$0.049) & 0.087~($\pm$0.045) & 35.52~($\pm$1.26) \\
			& & {\methodsp (SLM-FP)} & 0.039~($\pm$0.031) & 0.087~($\pm$0.016) & \textbf{0.048}~($\pm$0.024) & \textbf{38.17}~($\pm$0.96) \\
			
			\midrule
			
			\multirow{6}{*}{ID} & \multirow{6}{*}{SemEval2018} & {AA (IG)} & -0.120~($\pm$0.061) & 0.128~($\pm$0.014) & 0.248~($\pm$0.064) & 29.63~($\pm$4.72) \\
			& & {AA-F (Rand)} & -0.133~($\pm$0.043) & 0.124~($\pm$0.013) & 0.258~($\pm$0.053) & 32.39~($\pm$9.73) \\
			& & {\methodsp (AA-F)} & -0.028~($\pm$0.051) & 0.069~($\pm$0.041) & 0.096~($\pm$0.011) & \textbf{49.95}~($\pm$8.31) \\
			& & {\methodsp (DLM-P)} & -0.112~($\pm$0.095) & 0.140~($\pm$0.017) & 0.252~($\pm$0.112) & 27.78~($\pm$5.08) \\
			& & {\methodsp (DLM-FP)} & \textbf{0.047}~($\pm$0.017) & \textbf{0.149}~($\pm$0.052) & 0.102~($\pm$0.053) & 31.97~($\pm$2.80) \\
			& & {\methodsp (SLM-FP)} & 0.027~($\pm$0.047) & 0.091~($\pm$0.027) & \textbf{0.064}~($\pm$0.033) & 17.42~($\pm$4.04) \\
			
			\bottomrule 
			
		\end{tabular}
	}
	\vspace{0.3cm}
	\caption{\small \textbf{Zero-Shot Faithfulness Transfer from SST.}
		We investigate whether the faithfulness of \methodsp rationale extractors (AA-F, DLM-FP) trained on SST can generalize to unseen datasets/tasks, even when the task model's task performance cannot.
		Also, we include AA (IG) as a heuristic extractor baseline (\ie only the task model is trained).
		Here, the seen task is sentiment analysis (SA), while the unseen tasks are hate speech detection (HSD), offensive speech detection (OSD), and irony detection (ID).
		For SA, the unseen datasets are Yelp and Amazon.
		For HSD, OSD, and ID, the unseen datasets are Stormfront, OffenseEval, and SemEval2018, respectively.
		Overall, we find that faithfulness is not strongly correlated with task performance, as unseen tasks' comp/suff scores are similar to seen tasks'.
		In particular, though all methods achieve poor task performance on unseen tasks, \methodsp (DLM-FP)'s comp/suff scores are consistently good across all tasks, demonstrating its faithfulness generalization ability.
	}
	\label{tab:transfer}
\end{table*}

\subsection{Zero-Shot Faithfulness Transfer}
\label{sec:experiments:zs}

In Table \ref{tab:transfer}, we investigate if $\mathcal{F}_{\text{ext}}$'s faithfulness, obtained via UNIREX training on some source (seen) dataset, can generalize to target (unseen) datasets/tasks in a zero-shot setting (\ie no fine-tuning on target datasets/tasks).
In this experiment, we consider SST and sentiment analysis as the source dataset and task, respectively.
We compare six methods: AA (IG), AA-F (Rand), \methodsp (AA-F), \methodsp (DLM-P), \methodsp (DLM-FP), and \methodsp (SLM-FP).
For AA (IG), only $\mathcal{F}_{\text{task}}$ is trained on SST, since its $\mathcal{F}_{\text{ext}}$ is heuristic.
Meanwhile, for the \methodsp variants, both $\mathcal{F}_{\text{task}}$ and $\mathcal{F}_{\text{ext}}$ are trained on SST.
First, as an in-domain reference point, we report faithfulness and task performance on SST.
Second, we evaluate on unseen target datasets for a seen task (\ie sentiment analysis): Yelp \citep{zhangCharacterlevelConvolutionalNetworks2015} and Amazon \citep{mcauley2013hidden}.
Third, we evaluate on unseen target datasets for unseen target tasks: Stormfront (hate speech detection, binary F1) \citep{de2018hate}, OffenseEval (offensive speech detection, macro F1) \citep{zampieri2019semeval}, and SemEval2018 (irony detection, binary F1) \citep{van2018semeval}.
For more details about these datasets, please refer to \textsection \ref{sec:appendix:datasets}.

We want to show that, even if $\mathcal{F}_{\text{task}}$ yields poor task performance on unseen datasets, $\mathcal{F}_{\text{ext}}$'s rationales can still be faithful.
To make the faithfulness results easier to digest, we introduce a metric called Comp-Suff Difference (CSD), which locally aggregates comp and suff as: $\text{CSD} = \text{comp} - \text{suff}$.
Therefore, since higher/lower comp/suff signals higher faithfulness, then higher CSD signals higher faithfulness.

As expected, all methods achieve much lower task performance in the third setting than in the first two settings.
However, faithfulness does not appear to be strongly correlated with task performance, as unseen tasks' comp/suff scores are similar to seen tasks'.
Across all datasets, DLM-FP has the best faithfulness and is the only method whose comp is always higher than suff.
The other \methodsp variants are not as consistently strong as DLM-FP, but almost always beat non-\methodsp methods on comp and suff.
Meanwhile, AA (IG) has the worst comp and suff overall.
Ultimately, these results suggest that \method-trained models' faithfulness (\ie alignment between $\mathcal{F}_{\text{task}}$'s and $\mathcal{F}_{\text{ext}}$'s outputs) is a dataset/task agnostic property (\ie can generalize across datasets/tasks), further establishing \method's utility in low-resource settings.



\subsection{Plausibility User Study}
\label{sec:experiments:user_study}

Gold rationale based plausibility evaluation is noisy because gold rationales are for the target label, not a $\mathcal{F}_{\text{task}}$'s predicted label.
Thus, we conduct two five-annotator user studies (Table \ref{tab:user_studies}) to get a better plausibility measurement.
Given 50 random test instances from SST, we get the rationales for SGT+P, A2R+P, \methodsp (AA-FP), and \methodsp (DLM-FP), plus the gold rationales.
For each instance, we threshold all rationales to have the same number of positive tokens as the gold rationale.
The first user study is forward simulation \citep{hase2020evaluating, jain2020learning}.
Here, the annotator is given an input and a rationale for some model's prediction, then asked what (binary) sentiment label the model most likely predicted.
For forward simulation, we also consider a No Rationale baseline, where no tokens are highlighted.
For No Rationale and Gold (which we call ``oracle methods''), the target label is the correct choice.
Annotators are also asked to rate their confidence (4-point Likert scale) in their answer to this question. 
The second user study involves giving a subjective rating of how plausible the rationale is \citep{hase2020evaluating}.
Here, the annotator is given the input, rationale, and model's predicted label, then asked to rate (5-point Likert scale) how aligned the rationale is with the prediction.

In both accuracy and subjective rating, we find that DLM-FP performs best among all non-oracle methods and even slightly beats Gold on accuracy, further supporting our claim that DLM-FP rationales are plausible.
As expected, the fact that Gold does not achieve near-100$\%$ accuracy shows the discrepancy between evaluating plausibility based on the target label (\ie gold rationale similarity) and $\mathcal{F}_{\text{task}}$'s predicted label (forward simulation).
Meanwhile, SGT+P and AA-FP, which had lower AUPRC and TF1 in our automatic evaluation, also do worse in accuracy and alignment.
Also, users found SGT+P and AA-FP rationales harder to understand, as shown by their lower confidence scores.
Meanwhile, A2R+P had high AUPRC and TF1, but gets very low accuracy and alignment because A2R+P's predicted label was often not the target label, leading to misalignment with its gold-like rationale.
Nonetheless, users were still most confident in their predictions using A2R+P's rationales.
A2R+P is a great example of how automatic plausibility evaluation can be misleading.
For the accuracy, confidence, and alignment questions, we achieved Fleiss' Kappa \citep{fleiss1971measuring} inter-annotator agreement scores of 0.2456 (fair), 0.1282 (slight), and, 0.1561 (slight), respectively.
This lack of agreement demonstrates the difficulty of measuring rationale plausibility.

\begin{table}[ht!]
	\centering
	\scalebox{0.70}{
		\begin{tabular}{cccc}
			\toprule
			\multirow{2}{*}{\textbf{Method}} & \multicolumn{2}{c}{\textbf{Forward Simulation}} & \textbf{Subjective Rating} \\
			\cmidrule(lr){2-3} \cmidrule(lr){4-4}
			& Accuracy ($\%$) & Confidence (1-4) & Alignment (1-5) \\
			\midrule
			{No Rationale} & 92.00~($\pm$3.35) & 3.02~($\pm$0.39) & - \\
			
			\midrule
			
			{SGT+P} & 80.80~($\pm$9.73) & 2.34~($\pm$0.31) & 3.64~($\pm$0.28) \\
			
			{A2R+P} & 41.20~($\pm$4.71) & \textbf{2.83}~($\pm$0.28) & 2.97~($\pm$0.12) \\
			
			{\methodsp (AA-FP)} & 72.00~($\pm$7.78) & 2.00~($\pm$0.31) & 3.26~($\pm$0.31) \\
			
			{\methodsp (DLM-FP)} & \textbf{83.60}~($\pm$5.41) & 2.77~($\pm$0.28) & \textbf{3.96}~($\pm$0.22) \\
			
			\midrule
			
			{Gold} & 81.20~($\pm$3.03) & 2.88~($\pm$0.30) & 4.00~($\pm$0.20) \\
			
			\bottomrule 
		\end{tabular}
	}
	\caption{\small \textbf{Plausibility User Study on SST.} In our user study, we find that humans judge \methodsp (DLM-FP)'s rationales as being more plausible than those created by methods, in terms of both forward simulation (accuracy) and subjective rating (alignment).}
	\label{tab:user_studies}
	\vspace{-0.1cm}
\end{table}

\section{Related Work}
\label{sec:related_work}



\textbf{Extractive Rationale Faithfulness~}
Many prior works have tried to improve the faithfulness of extractive rationales through the use of AAs \citep{bastings2020elephant}.
Typically, this involves handcrafting gradient-based \citep{sundararajan2017axiomatic, denil2014extraction, lundberg2017unified, li2015visualizing} or perturbation-based \citep{li2016understanding, poerner2018evaluating, kadar2017representation} AAs.  
However, attribution algorithms cannot be optimized and tend to be compute-intensive (often requiring multiple LM forward/backward passes).
Recently, \citet{ismail2021improving} addressed the optimization issue by regularizing the task model to yield faithful rationales via the AA, while other works \citep{situ2021learning, schwarzenberg2021efficient} addressed the compute cost issue by training an LM (requiring only one forward pass) to mimic an AA's behavior.
Another line of work aims to produce faithful rationales by construction, via SPPs \citep{jain2020learning, yu2021understanding, paranjape2020information, bastings2019interpretable, yu2019rethinking, lei2016rationalizing}.
Still, SPPs' faithfulness can only guarantee sufficiency -- not comprehensiveness \citep{deyoung2019eraser}.
Also, SPPs generally perform worse than vanilla LMs because they hide much of the original text input from the predictor and are hard to train end-to-end.


\textbf{Extractive Rationale Plausibility~}
Existing approaches for improving extractive rationale plausibility typically involve supervising LM-based extractors \citep{bhat2021self} or SPPs \citep{jain2020learning, paranjape2020information, deyoung2019eraser} with gold rationales.
However, existing LM-based extractors have not been trained for faithfulness, while SPPs' faithfulness by construction comes at the great cost of task performance.
Meanwhile, more existing works focus on improving the plausibility of free-text rationales \citep{narang2020wt5, lakhotia2020fid, camburu2018snli}, often with task-specific pipelines \citep{rajani2019explain, kumar2020nile}.

\textbf{Connection to \method~}
Unlike prior works, \methodsp enables both the task model and rationale extractor to be jointly optimized for faithfulness, plausibility, and task performance.
As a result, \method-trained rationale extractors achieve a better balance of faithfulness and plausibility, without compromising the task model's performance.
Also, by using a learned rationale extractor, which generally only requires one model forward pass, \methodsp does not have the computational expenses that limit many AAs.

\bibliography{references}
\bibliographystyle{icml2022}

\newpage
\appendix
\section{Appendix} 
\label{sec:appendix}

\subsection{Text Classification}
\label{sec:appendix:text_cls}
Here, we formalize the text classification problem in more detail.
Let $\mathcal{D} = \{\mathcal{X}, \mathcal{Y}\}_{i=1}^{N}$ be a dataset, where $\mathcal{X} = \{\mathbf{x}_i\}_{i=1}^{N}$ are the text inputs, $\mathcal{Y} = \{y_{i}^{*}\}_{i=1}^{N}$ are the labels, and $N$ is the number of instances $(\mathbf{x}_i, y_{i}^{*})$ in $\mathcal{D}$.
We also assume $\mathcal{D}$ can be partitioned into train set $\mathcal{D}_{\text{train}}$, dev set $\mathcal{D}_{\text{dev}}$, and test set $\mathcal{D}_{\text{test}}$.
Let $\mathcal{F}_{\text{task}} = f_{\text{task}}(f_{\text{enc}}(\cdot))$ be a task LM, where $f_{\text{enc}}$ is the text encoder, and $f_{\text{task}}$ is the task output head.
Typically, $\mathcal{F}_{\text{task}}$ has a BERT-style architecture \citep{devlin2018bert}, in which $f_{\text{enc}}$ is a Transformer \citep{vaswani2017attention} while $f_{\text{task}}$ is a linear layer.
Below, we define the sequence classification (SST, Movies, MultiRC, e-SNLI) and multi-choice QA (CoS-E) tasks, which are different types of text classification.

\paragraph{Sequence Classification}
In sequence classification, $\mathbf{x}_i$ is a token sequence (\eg a single sentence, a pair of sentences), while $y_{i}^{*}$ is the target class for $\mathbf{x}_i$.
Here, we assume a fixed label space $Y = \{1, ... , M\}$ of size $M$, where $y_{i}^{*} \in Y$ for all $i$.
Thus, $f_{\text{task}}$ outputs a vector of size $M$, such that $\mathcal{F}_{\text{task}}(\mathbf{x}_i) = f_{\text{task}}(f_{\text{enc}}(\mathbf{x}_i)) = \hat{\mathbf{y}}_i \in \R^{M}$ is the logit vector used to classify $\mathbf{x}_i$.
Given $\hat{\mathbf{y}}_i = [\hat{y}_{i,j}]_{j=1}^{M}$, let $y_i = \argmax_{\hspace{0.5mm} j} \hat{y}_{i,j}$ be the class predicted by $\mathcal{F}_{\text{task}}$.
The goal of sequence classification is to learn $\mathcal{F}_{\text{task}}$ such that $y_{i}^{*} = y_i$, for all $(\mathbf{x}_i, y_{i}^{*})$ \citep{minaee2021deep}.

\paragraph{Multi-Choice QA}
Instead of a fixed label space, multi-choice QA has a different (but fixed-size) set of answer choices per instance.
For instance $i$, let $q_i$ be the question (\eg \textit{``A friend is greeting me, what would they say?''}) and $A_i = \{a_{i,j}\}_{j=1}^{M}$ be the corresponding answer choices (\eg \{\textit{``say hello''}, \textit{``greet''}, \textit{``associate''}, \textit{``socialize''}, \textit{``smile''}\}), where $M$ is now the number of answer choices.
Define $\mathbf{x}_{i,j} = q_i \oplus a_{i,j}$, where $\oplus$ denotes concatenation.
In multi-choice QA, we have $\mathbf{x}_i = \{\mathbf{x}_{i,j}\}_{j=1}^{M}$, while $y_{i}^{*} \in A_i$ is the correct answer for $\mathbf{x}_i$.
Thus, $f_{\text{task}}$ outputs a scalar, such that $\mathcal{F}_{\text{task}}(\mathbf{x}_{i,j}) = f_{\text{task}}(f_{\text{enc}}(\mathbf{x}_{i,j})) = \hat{y}_{i,j} \in \R$ is the logit for $\mathbf{x}_{i,j}$.
Given $\hat{\mathbf{y}}_i = [\hat{y}_{i,j}]_{j=1}^{M}$, let $j' = \argmax_{\hspace{0.5mm} j} \hat{y}_{i,j}$, where $y_i = a_{i,j'}$ is the answer predicted by $\mathcal{F}_{\text{task}}$.
The goal of multi-choice QA is to learn $\mathcal{F}_{\text{task}}$ such that $y_{i}^{*} = y_i$, for all $(\mathbf{x}_i, y_{i}^{*})$ \citep{talmor2018commonsenseqa}.

\subsection{Gold Rationale Supervision}
\label{sec:appendix:gold}

If a learned rationale extractor is chosen, \methodsp enables users to specify how much gold rationale supervision to use.
Ideally, each train instance would be annotated with a gold rationale.
In this case, we could directly minimize the plausibility loss for each train instance.
However, since gold rationales can be expensive to annotate, \methodsp provides a special batching procedure for training with limited gold rationale supervision.

Given $N_{\text{train}} = |\mathcal{D}_{\text{train}}|$ train instances, let $0 < \gamma < 100$ be the percentage of train instances with gold rationales, $N_{\text{gold}} = \ceil{\frac{\gamma}{100} N_{\text{train}}} \geq 1$ be the number of train instances with gold rationales, $b$ be the desired train batch size, and $\beta > 1$ be a scaling factor.
Define $\mathcal{D}_{\text{gold}} \subseteq \mathcal{D}_{\text{train}}$ as the set of train instances with gold rationales, where $|\mathcal{D}_{\text{gold}}| = N_{\text{gold}}$.
Note that, if all train instances have gold rationales, then $\mathcal{D}_{\text{gold}} = \mathcal{D}_{\text{train}}$ and $\gamma = 100$.

Each batch is constructed as follows: (1) randomly sample $b_{\text{gold}} = \max(1, \frac{b}{\beta})$ instances from $\mathcal{D}_{\text{gold}}$ without replacement, then (2) randomly sample $b - b_{\text{gold}}$ instances from $\mathcal{D}_{\text{train}} \backslash \mathcal{D}_{\text{gold}}$ without replacement.
This results in a batch with $b$ total train instances, $b_{\text{gold}}$ with gold rationales and the rest without.
Since $N_{\text{gold}}$ is generally small, we only sample from $\mathcal{D}_{\text{gold}}$ without replacement for a given batch, but not a given epoch.
Thus, instances from $\mathcal{D}_{\text{gold}}$ may appear more than once in the same epoch.
However, we do sample from $\mathcal{D}_{\text{train}} \backslash \mathcal{D}_{\text{gold}}$ without replacement for each batch and epoch, so every instance in $\mathcal{D}_{\text{train}} \backslash \mathcal{D}_{\text{gold}}$ appears exactly once per epoch.

After constructing the batch, we compute the plausibility loss for the batch as follows: $\sum_{i=1}^{b} \mathbbm{1}_{(\textbf{x}_i, y_{i}^{*}) \in \mathcal{D}_{\text{gold}}}$ $\mathcal{L}_{\text{plaus}}(\mathcal{F}_{\text{ext}}(\textbf{x}_i), \textbf{r}_{i}^{*})$, where $\mathcal{L}_{\text{plaus}}$ is the plausibility loss for train instance $(\textbf{x}_i, y_{i}^{*})$.
This function zeroes out the plausibility loss for instances without gold rationales, so that plausibility is only being optimized with respect to instances with gold rationales.
However, in \textsection \ref{sec:experiments:gold_eff}, we show that it is possible to achieve high plausibility via rationale extractors trained on minimal gold rationale supervision.

\subsection{Explainability Objectives}
\label{sec:appendix:expl_objectives}

\subsubsection{Faithfulness}
\label{sec:appendix:expl_objectives:faithfulness}

\paragraph{Sufficiency}


In addition, to the criteria presented in \textsection \ref{sec:method:expl_objectives}, we consider two other sufficiency loss functions.
The first is the \textit{KL divergence criterion} used in \citep{ismail2021improving}, which considers the entire label distribution and is defined as $\mathcal{L}_{\text{suff-KL}} = \text{KL}(\mathcal{F}_{\text{task}}(\mathbf{r}_{i}^{(k)})) \hspace{1mm} || \hspace{1mm} \mathcal{F}_{\text{task}}(\textbf{x}_i))$.
The second is the \textit{mean absolute error (MAE) criterion}, which is defined as $\mathcal{L}_{\text{suff-MAE}} = | \mathcal{L}_{\text{CE}}(\mathcal{F}_{\text{task}}(\mathbf{r}_{i}^{(k)})), y_{i}^{*}) - \mathcal{L}_{\text{CE}}(\mathcal{F}_{\text{task}}(\textbf{x}_i), y_{i}^{*})|$.
Unlike the difference criterion $\mathcal{L}_{\text{suff-diff}}$ and margin criterion $\mathcal{L}_{\text{suff-margin}}$ (\textsection \ref{sec:method:expl_objectives}), the MAE criterion assumes that using $\mathbf{r}_{i}^{(k)}$ as input should not yield better task performance than using $\textbf{x}_i$ as input. 
In our experiments, we find that $\mathcal{L}_{\text{suff-margin}}$ is effective, though others (\eg KL divergence, MAE) can be used too.

\subsubsection{Plausibility}
\label{sec:appendix:expl_objectives:plausibility}
Similar to faithfulness, \methodsp places no restrictions on the choice of plausibility objective.
As described in \textsection \ref{sec:method:expl_objectives}, given gold rationale $\mathbf{r}_{i}^{*}$ for input $\mathbf{x}_i$, plausibility optimization entails training $\mathcal{F}_{\text{ext}}$ to predict binary importance label $\mathbf{r}_{i}^{*,t}$ for each token $x_i^t$.
This is essentially binary token classification, so one natural choice for $\mathcal{L}_{\text{plaus}}$ is the token-level binary cross-entropy (BCE) criterion: $\mathcal{L}_{\text{plaus-BCE}} = - \sum_t \mathbf{r}_{i}^{*,t}\log(\mathcal{F}_{\text{ext}}(x_i^t))$ (\textsection \ref{sec:method:expl_objectives}).
Another option is the sequence-level \textit{KL divergence criterion}, which is defined as: $\mathcal{L}_{\text{plaus-KL}} = \text{KL}(\mathcal{F}_{\text{ext}}(\textbf{x}_i) \hspace{1mm} || \hspace{1mm} \textbf{r}_{i}^{*})$.

Additionally, we can directly penalize $\mathcal{F}_{\text{ext}}(\textbf{x}_i)$ in the logit space via a \textit{linear loss}, defined as: $\mathcal{L}_{\text{plaus-linear}} = \Phi(\textbf{r}_{i}^{*}) \hspace{1mm} \mathcal{F}_{\text{ext}}(\textbf{x}_i)$, where $\Phi(u) = -2u + 1$ maps positive and negative tokens to $-1$ and $+1$, respectively. The linear loss directly pushes the logits corresponding to positive/negative tokens to be higher/lower and increase the margin between them. To prevent linear loss values from becoming arbitrarily negative, we can also lower bound the loss with a margin $m_p$, yielding: $\mathcal{L}_{\text{plaus-linear-margin}} = \max(-m_p, \mathcal{L}_{\text{plaus-linear}}) + m_p$.

\subsection{Datasets}
\label{sec:appendix:datasets}

As described in \textsection \ref{sec:experiments:eval:datasets}, we primarily experiment with the SST (sentiment analysis) \citep{socher2013recursive, carton2020evaluating}, Movies (sentiment analysis) \citep{zaidan2008modeling}, CoS-E (commonsense question answering) \citep{rajani2019explain}, MultiRC (reading comprehension) \citep{khashabi2018looking}, and e-SNLI (natural language inference) \citep{camburu2018snli} datasets, all of which have gold rationale annotations.
The rationale-annotated version of SST was obtained from \citet{carton2020evaluating}, while the latter four datasets were obtained from the ERASER benchmark \citep{deyoung2019eraser}.
The numbers of train/dev/test instances in these datasets are as follows: SST (6920/872/1821), Movies (1599/200/200), CoS-E (8752/1086/1079), MultiRC (24029/3214/4848), and e-SNLI (549309/9823/9807).

\begin{table*}[ht!]
	\centering
	\scalebox{0.68}{
		\begin{tabular}{ccccccccc}
			\toprule
			\multirow{2}{*}{\textbf{Method}} & \multicolumn{3}{c}{\textbf{Faithfulness}} & \multicolumn{2}{c}{\textbf{Plausibility}} & \multicolumn{1}{c}{\textbf{Task Perf.}} &
			\multicolumn{1}{c}{\textbf{IG}} &
			\multicolumn{1}{c}{\textbf{Comp. Efficiency}} \\
			\cmidrule(lr){2-4} \cmidrule(lr){5-6} \cmidrule(lr){7-7} \cmidrule(lr){8-8} \cmidrule(lr){9-9}
			& CSD ($\uparrow$) & Comp ($\uparrow$) & Suff ($\downarrow$) & AUPRC ($\uparrow$) & TF1 ($\uparrow$) & Acc ($\uparrow$) & Conv. Delta ($\downarrow$) & Inference Time ($\downarrow$) \\
			
			\midrule
			
			{AA (IG-3)} & -0.138~($\pm$0.040) & 0.119~($\pm$0.009) & 0.258~($\pm$0.031) & 49.94~($\pm$1.77) & 50.75~($\pm$0.54) & \textbf{93.81}~($\pm$0.55) & 8.07~($\pm$1.47) & \textbf{7.94E-03}~($\pm$5.30E-05) \\
			
			{AA (IG-5)} & -0.141~($\pm$0.031) & 0.134~($\pm$0.015) & 0.275~($\pm$0.021) & 49.38~($\pm$1.00) & 50.85~($\pm$0.70) & \textbf{93.81}~($\pm$0.55) & 6.83~($\pm$0.92) & 1.15E-02~($\pm$6.03E-05) \\
			
			{AA (IG-10)} & 0.011~($\pm$0.043) & 0.222~($\pm$0.015) & 0.210~($\pm$0.031) & 55.87~($\pm$0.51) & 52.06~($\pm$0.33) & \textbf{93.81}~($\pm$0.55) & 6.55~($\pm$0.48) & 2.08E-02~($\pm$1.14E-04) \\
			
			{AA (IG-30)} & 0.056~($\pm$0.058) & 0.258~($\pm$0.020) & 0.202~($\pm$0.038) & 57.23~($\pm$1.16) & 52.74~($\pm$0.43) & \textbf{93.81}~($\pm$0.55) & 4.91~($\pm$1.82) & 5.80E-02~($\pm$3.03E-04) \\
			
			{AA (IG-50)} & 0.066~($\pm$0.057) & 0.265~($\pm$0.017) & 0.199~($\pm$0.040) & 57.70~($\pm$1.02) & 52.66~($\pm$0.37) & \textbf{93.81}~($\pm$0.55) & 2.89~($\pm$0.34) & 9.54E-02~($\pm$5.64E-04) \\
			
			{AA (IG-70)} & 0.072~($\pm$0.055) & 0.269~($\pm$0.016) & 0.197~($\pm$0.039) & 58.11~($\pm$1.21) & 52.96~($\pm$0.32) & \textbf{93.81}~($\pm$0.55) & 2.50~($\pm$0.25) & 1.33E-01~($\pm$8.10E-04) \\
			
			{AA (IG-100)} & 0.074~($\pm$0.055) & 0.271~($\pm$0.016) & 0.197~($\pm$0.039) & 58.25~($\pm$1.27) & 53.00~($\pm$0.42) & \textbf{93.81}~($\pm$0.55) & 2.01~($\pm$0.13) & 1.89E-01~($\pm$1.62E-03) \\
			
			{AA (IG-500)} & 0.082~($\pm$0.055) & 0.276~($\pm$0.016) & 0.195~($\pm$0.039) & 58.61~($\pm$1.10) & \textbf{53.25}~($\pm$0.29) & \textbf{93.81}~($\pm$0.55) & 0.99~($\pm$0.23) & 9.38E-01~($\pm$5.44E-03) \\
			
			{AA (IG-1000)} & 0.083~($\pm$0.057) & 0.278~($\pm$0.017) & 0.195~($\pm$0.040) & \textbf{58.64}~($\pm$1.15) & 53.17~($\pm$0.39) & \textbf{93.81}~($\pm$0.55) & \textbf{0.74}~($\pm$0.16) & 1.88E+00~($\pm$1.67E-03) \\
			
			{\methodsp (AA-F)} & \textbf{0.120}~($\pm$0.055) & \textbf{0.292}~($\pm$0.051) & \textbf{0.171}~($\pm$0.038) & 48.13~($\pm$1.14) & 50.96~($\pm$0.93) & 92.97~($\pm$0.44) & 8.07~($\pm$1.47) & \textbf{7.94E-03}~($\pm$5.30E-05) \\
			
			\midrule
			
			{\methodsp (DLM-FP)} & 0.151~($\pm$0.056) & \textbf{0.319}~($\pm$0.090) & 0.167~($\pm$0.036) & \textbf{85.80}~($\pm$0.74) & \textbf{72.76}~($\pm$0.19) & \textbf{93.81}~($\pm$0.54) & - & \textbf{8.51E-04}~($\pm$7.82E-07) \\
			
			{\methodsp (SLM-FP)} & \textbf{0.189}~($\pm$0.030) & 0.302~($\pm$0.039) & \textbf{0.113}~($\pm$0.013) & 82.55~($\pm$0.84) & 70.65~($\pm$0.44) & 93.68~($\pm$0.67) & - & 8.81E-04~($\pm$5.67E-06) \\
			
			\bottomrule 
		\end{tabular}
	}
	\caption{\small{\textbf{Computational Efficiency Results on SST.}} Besides achieving the best balance of faithfulness, plausibility, and task performance, \methodsp (DLM-FP) and \methodsp (SLM-FP) are also the most computationally efficient, achieving the lowest inference time per instance. Also, despite having a higher convergence delta and lower inference time than most other AA (IG) variants due to using 3-step IG, \methodsp (AA-F) outperforms all AA (IG) variants on faithfulness, while achieving comparable plausibility and task performance.}
	\label{tab:compute}
\end{table*}

As described in \textsection \ref{sec:experiments:zs}, for our zero-shot experiments, we consider the Yelp (sentiment analysis) \citep{zhangCharacterlevelConvolutionalNetworks2015}, Amazon (sentiment analysis) \citep{mcauley2013hidden}, Stormfront (hate speech detection) \citep{de2018hate}, OffenseEval (offensive speech) \citep{zampieri2019semeval}, and SemEval2018 (irony detection) \citep{van2018semeval} datasets.
The numbers of train/dev/test instances in these datasets are as follows: Yelp (10000/2000/2000), Amazon (10000/2000/2000), Stormfront (7896/978/1998), OffenseEval (11916/1324/860), and SemEval2018 (2862/955/784).
For Yelp and Amazon, the original datasets are very large, so we use stratified sampling with respect to labels to obtain the final train/dev/test split split of 10000/2000/2000.



\begin{figure}[ht!]
	\includegraphics[width=0.48\textwidth]{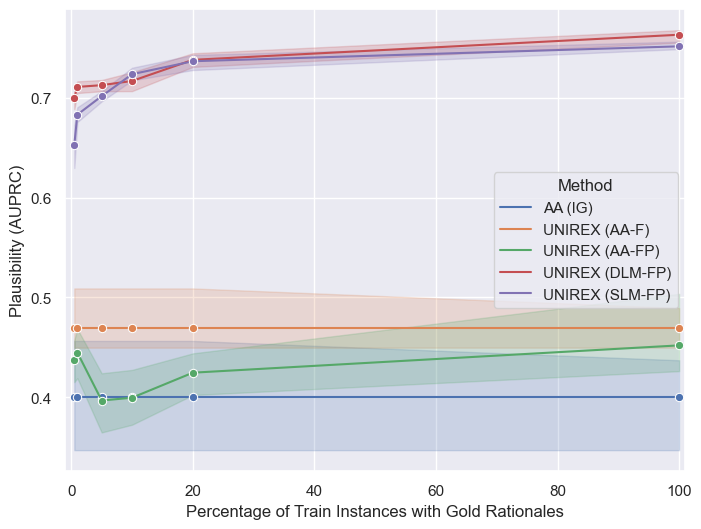}
	\caption{\small \textbf{Gold Rationale Efficiency on CoS-E.} As shown in this plot, \methodsp (DLM-FP) and \methodsp (SLM-FP) are able to achieve high plausibility performance, even with a very small percentage of training instances with gold rationale annotations.}
	\label{fig:gold_eff:cose}
\end{figure} 

\subsection{Gold Rationale Efficiency}
\label{sec:appendix:gold_eff}
Fig. \ref{fig:gold_eff:cose} shows the gold rationale data efficiency results for CoS-E, using the same setup as \textsection \ref{sec:experiments:gold_eff}.
Overall, we see that the CoS-E results are quite similar to the SST results.
Again, \methodsp (DLM-FP) and \methodsp (SLM-FP) dominate across all $\gamma$ values, with AUPRC slowly decreasing as $\gamma$ decreases.
Interestingly, \methodsp (AA-FP) yields a noticeable dip in AUPRC for lower $\gamma$ values.
Since AA-FP has limited capacity (via the task model) for plausibility optimization, it is possible that this fluctuation is due to random noise.
We leave further analysis of this for future work.

\subsection{Computational Efficiency}
\label{sec:appendix:compute}
Besides faithfulness, plausibility, and task performance, computational efficiency is also an important desideratum of rationale extraction.
With that in mind, the number of IG steps is a critical design choice.
A higher number of IG steps means a more accurate IG approximation, which should may yield more faithful rationales.
On the other hand, a higher number of IG steps means greater computational costs.
As stated in \textsection \ref{sec:experiments:implementation}, we use 3-step IG in all of our experiments, in order to make all methods computationally comparable.
3-step IG requires three backward passes, while all other compared methods require either one forward pass or one backward pass.
We would like to empirically characterize this trade-off between faithfulness and computational efficiency.
Thus, we compare IG performance for $\eta = [3, 5, 10, 30, 50, 70, 100, 500, 1000]$ steps, denoted as AA (IG-$\eta$).
In Table \ref{tab:compute}, we report faithfulness, plausibility, task performance, convergence delta (\ie IG approximation error), and inference time for each IG-based method.
Additionally, we compare these IG settings to \methodsp (AA-F) (which uses 3-step IG), \methodsp (DLM-FP), and \methodsp (SLM-FP).
Since \methodsp (DLM-FP) and \methodsp (SLM-FP) are not IG-based, we do not report convergence delta for them.

\begin{table*}[ht!]
\centering
\scalebox{0.68}{
\begin{tabular}{cccccc}
    \toprule  
    \textbf{Instance Ranking} & \textbf{Instance ID} & \textbf{Method} & \textbf{Rationale} & $\mathcal{F}_{\text{task}}$\textbf{'s Prediction} & \textbf{Alignment (1-5)} \\
    
    \midrule
    
    \multirow{15}{*}{Top Alignment Mean} & \multirow{5}{*}{30} & {SGT+P} & {\color{blue}{\textbf{good}}} {\color{blue}{\textbf{actress}}} . & positive & 5.00~($\pm$0.00) \\
    & & {A2R+P} & {\color{blue}{\textbf{good}}} {\color{blue}{\textbf{actress}}} . & positive & 5.00~($\pm$0.00) \\
    & & {\methodsp (AA-FP)} & {\color{blue}{\textbf{good}}} actress {\color{blue}{\textbf{.}}} & positive & 4.80~($\pm$0.45) \\
    & & {\methodsp (DLM-FP)} & {\color{blue}{\textbf{good}}} {\color{blue}{\textbf{actress}}} . & positive & 5.00~($\pm$0.00) \\
    & & {Gold} & {\color{blue}{\textbf{good}}} {\color{blue}{\textbf{actress}}} . & positive & 5.00~($\pm$0.00) \\
    
    \cmidrule(lr){2-6}
    
    & \multirow{5}{*}{41} & {SGT+P} & a cop story {\color{blue}{\textbf{that}}} {\color{blue}{\textbf{understands}}} {\color{blue}{\textbf{the}}} {\color{blue}{\textbf{medium}}} {\color{blue}{\textbf{amazingly}}} {\color{blue}{\textbf{well}}} . & positive & 5.00~($\pm$0.00) \\
    & & {A2R+P} & {\color{blue}{\textbf{a}}} {\color{blue}{\textbf{cop}}} story that {\color{blue}{\textbf{understands}}} the {\color{blue}{\textbf{medium}}} {\color{blue}{\textbf{amazingly}}} {\color{blue}{\textbf{well}}} . & positive & 4.80~($\pm$0.45) \\
    & & {\methodsp (AA-FP)} & {\color{blue}{\textbf{a}}} {\color{blue}{\textbf{cop}}} {\color{blue}{\textbf{story}}} that {\color{blue}{\textbf{understands}}} the medium {\color{blue}{\textbf{amazingly}}} well {\color{blue}{\textbf{.}}} & positive & 4.20~($\pm$0.45) \\
    & & {\methodsp (DLM-FP)} & a {\color{blue}{\textbf{cop}}} story that {\color{blue}{\textbf{understands}}} {\color{blue}{\textbf{the}}} {\color{blue}{\textbf{medium}}} {\color{blue}{\textbf{amazingly}}} {\color{blue}{\textbf{well}}} . & positive & 4.80~($\pm$0.45) \\
    & & {Gold} & a cop story that {\color{blue}{\textbf{understands}}} {\color{blue}{\textbf{the}}} {\color{blue}{\textbf{medium}}} {\color{blue}{\textbf{amazingly}}} {\color{blue}{\textbf{well}}} {\color{blue}{\textbf{.}}} & positive & 5.00~($\pm$0.00) \\
    
    \cmidrule(lr){2-6}
    
    & \multirow{5}{*}{35} & {SGT+P} & chicago is , in many {\color{blue}{\textbf{ways}}} , an {\color{blue}{\textbf{admirable}}} achievement . & positive & 4.60~($\pm$0.55) \\
    & & {A2R+P} & chicago is , in many ways , an {\color{blue}{\textbf{admirable}}} {\color{blue}{\textbf{achievement}}} . & positive & 5.00~($\pm$0.00) \\
    & & {\methodsp (AA-FP)} & {\color{blue}{\textbf{ch}}}icago is , in many {\color{blue}{\textbf{ways}}} , an admirable achievement . & positive & 3.20~($\pm$0.45) \\
    & & {\methodsp (DLM-FP)} & chicago is , in many ways , an {\color{blue}{\textbf{admirable}}} {\color{blue}{\textbf{achievement}}} . & positive & 5.00~($\pm$0.00) \\
    & & {Gold} & chicago is , in many ways , an {\color{blue}{\textbf{admirable}}} {\color{blue}{\textbf{achievement}}} . & positive & 5.00~($\pm$0.00) \\
    
    \midrule
    
    \multirow{15}{*}{Top Alignment Std} & \multirow{5}{*}{9} & {SGT+P} & {\color{blue}{\textbf{bad}}} beyond belief and ridiculous beyond description {\color{blue}{\textbf{.}}} & negative & 4.60~($\pm$0.55) \\
    & & {A2R+P} & {\color{blue}{\textbf{bad}}} beyond belief and {\color{blue}{\textbf{ridiculous}}} beyond description . & positive & 1.00~($\pm$0.00) \\
    & & {\methodsp (AA-FP)} & bad beyond belief {\color{blue}{\textbf{and}}} ridiculous beyond description {\color{blue}{\textbf{.}}} & negative & 2.60~($\pm$0.89) \\
    & & {\methodsp (DLM-FP)} & {\color{blue}{\textbf{bad}}} beyond belief and {\color{blue}{\textbf{ridiculous}}} beyond description . & negative & 4.80~($\pm$0.45) \\
    & & {Gold} & {\color{blue}{\textbf{bad}}} beyond belief and {\color{blue}{\textbf{ridiculous}}} beyond description . & negative & 4.80~($\pm$0.45) \\
    
    \cmidrule(lr){2-6}
    
    & \multirow{5}{*}{12} & {SGT+P} & {\color{blue}{\textbf{these}}} are names to remember , in order to avoid them in the future . & negative & 2.40~($\pm$0.89) \\
    & & {A2R+P} & these are names to remember , in order to {\color{blue}{\textbf{avoid}}} them in the future . & positive & 1.20~($\pm$0.45) \\
    & & {\methodsp (AA-FP)} &  these are names to remember , in order to avoid them in the future {\color{blue}{\textbf{.}}} & negative & 2.60~($\pm$0.89) \\
    & & {\methodsp (DLM-FP)} & these are names to remember , in order to {\color{blue}{\textbf{avoid}}} them in the future . & negative & 4.80~($\pm$0.45) \\
    & & {Gold} & these are names to remember , in order to {\color{blue}{\textbf{avoid}}} them in the future . & negative & 4.80~($\pm$0.45) \\
    
    \cmidrule(lr){2-6}
    
    & \multirow{5}{*}{19} & {SGT+P} & the title {\color{blue}{\textbf{'s}}} lameness should clue you {\color{blue}{\textbf{in}}} on how {\color{blue}{\textbf{bad}}} {\color{blue}{\textbf{the}}} {\color{blue}{\textbf{movie}}} is . & negative & 4.00~($\pm$0.00) \\
    & & {A2R+P} & the title 's {\color{blue}{\textbf{lam}}}{\color{blue}{\textbf{eness}}} should {\color{blue}{\textbf{clue}}} {\color{blue}{\textbf{you}}} in on how {\color{blue}{\textbf{bad}}} the {\color{blue}{\textbf{movie}}} is . & positive & 1.20~($\pm$0.45) \\
    & & {\methodsp (AA-FP)} & the title {\color{blue}{\textbf{'s}}} {\color{blue}{\textbf{lam}}}eness {\color{blue}{\textbf{should}}} clue you {\color{blue}{\textbf{in}}} {\color{blue}{\textbf{on}}} how bad {\color{blue}{\textbf{the}}} movie is {\color{blue}{\textbf{.}}} & negative & 2.60~($\pm$0.89) \\
    & & {\methodsp (DLM-FP)} & the title 's {\color{blue}{\textbf{lam}}}{\color{blue}{\textbf{eness}}} should {\color{blue}{\textbf{clue}}} {\color{blue}{\textbf{you}}} in on how {\color{blue}{\textbf{bad}}} {\color{blue}{\textbf{the}}} {\color{blue}{\textbf{movie}}} is . & negative & 4.80~($\pm$0.45) \\
    & & {Gold} & {\color{blue}{\textbf{the}}} {\color{blue}{\textbf{title}}} {\color{blue}{\textbf{'s}}} {\color{blue}{\textbf{lam}}}{\color{blue}{\textbf{eness}}} should clue you in on how {\color{blue}{\textbf{bad}}} the movie is . & negative & 4.80~($\pm$0.45) \\
    
    \bottomrule
\end{tabular}
}
\vspace{0.3cm}
\caption{\small \textbf{Qualitative Analysis on SST.} Building upon the quantitative results of our plausibility user study, our qualitative analysis further supports the notion that \methodsp (DLM-FP)'s rationales are more plausible than those created by other rationale extraction methods. In this table, we visualize each rationale by highlighting the important tokens (selected by the given method) in \textcolor{blue}{\textbf{blue}}.}
\label{tab:qual}
\end{table*}

As expected, we find that convergence delta decreases as $\eta$ increases.
Furthermore, AA (IG) faithfulness scores generally improve as $\eta$ increases, although the improvement begins to saturate at around $\eta = 50$.
Similarly, we find that AA (IG) plausibility scores for IG also tend to improve as $\eta$ increases, with the improvement also saturating around $\eta = 50$.
This makes sense because plausibility optimization involves regularizing the task model to yield rationales that are similar to gold rationales, but this regularization is less effective if the yielded rationales are less faithful to the task model.
However, despite these faithfulness and plausibility improvements, we see that inference time increases roughly linearly with respect to $\eta$.
In particular, AA (IG-1000) is over 200 times slower than AA (IG-3).
Meanwhile, despite only using 3-step IG, \methodsp (AA-F) beats all AA (IG) variants on faithfulness, while achieving comparable plausibility and task performance.
On the other hand, \methodsp (DLM-FP) and \methodsp (SLM-FP) achieve the best faithfulness, while also achieving the best plausibility and inference time by far, without sacrificing task performance.
Both \methodsp (DLM-FP) and \methodsp (SLM-FP) are over 2000 times faster than AA (IG-1000) and over nine times faster than AA (IG-3).
These results show that \methodsp (esp. using learned rationale extractor) is an effective framework for jointly optimizing rationale extraction for faithfulness, plausibility, and task performance, while also achieving high computational efficiency.

\subsection{Qualitative Analysis}
\label{sec:appendix:qual}

In \textsection \ref{sec:experiments:user_study}, we presented a user study of fifty SST test instances to further evaluate the plausibility of rationales produced by various methods.
To get additional insights about rationale plausibility, we conduct qualitative analysis of selected instances from the user study.
In Table \ref{tab:qual}, for each selected instance and a given method, we show the method's rationale, $\mathcal{F}_{\text{task}}$'s corresponding prediction, and the user-annotated alignment score for the rationale with respect to the prediction.
We consider two groups of selected instances.

First, we select the top-3 instances with respect to mean alignment score (across five annotators), averaged over all methods (\ie Instances 30, 41, and 35).
Not surprisingly, we find that $\mathcal{F}_{\text{task}}$'s prediction is very consistent across all methods.
For these three instances, all of the predicted labels happen to be both positive and correct (\ie same as gold label).
Similarly, the rationales are also rather consistent across methods.
In particular, A2R+P, \methodsp (DLM-FP), and Gold consistently yield the highest alignment scores.
In Instance 30, these three methods plausibly highlight ``good'' and ``actress''.
In Instance 41, these three methods plausibly highlight ``understands'', ``medium'', ``amazingly'', and ``well''.
In Instance 35, these three methods plausibly highlight ``admirable'' and ``achievement''.
Of course, this is expected for Gold, since gold rationales are human-annotated with respect to the gold label.
The high alignment scores for A2R+P and \methodsp (DLM-FP) also make sense since A2R+P and \methodsp (DLM-FP) yielded high PNRG (Fig. \ref{fig:main:desiderata}).

Second, we select the top-3 instances with respect to standard deviation (std) alignment score (across five annotators), averaged over all methods (\ie Instances 9, 12, 19).
This time, there is greater variance in the per-instance alignment scores across methods.
For these three instances, the negative label is predicted by all methods except A2R+P.
Like before, \methodsp (DLM-FP) and Gold consistently yield the highest alignment scores.
In Instance 9, these two methods plausibly highlight ``bad'' and ``ridiculous''.
In Instance 12, these two methods plausibly highlight ``avoid''.
In Instance 19, these two methods plausibly highlight ``lameness'' and ``bad''.
However, this time, A2R+P consistently yields the lowest alignment scores.
Even though A2R+P produces rationales that are similar to those of \methodsp (DLM-FP) and Gold (\ie aligning with the gold label), A2R+P's rationales do not support A2R+P's predicted label.
This illustrates the limitation of automatically evaluating plausibility via gold rationale similarity, as A2R+P achieved high PNRG.
Meanwhile, we see that \methodsp (DLM-FP) consistently yields the highest plausibility across various types of evaluation (\ie gold rationale similarity, forward simulation, subjective rating).



\subsection{Main Results (Extended)}

\label{sec:appendix:raw_results}

In \textsection \ref{sec:experiments}, Figs. \ref{fig:main:composite_1}-\ref{fig:main:desiderata} only reported the main results averaged over all datasets.
In this section, Tables \ref{tab:main:sst}-\ref{tab:main:esnli} provide more detailed main results by reporting all raw and NRG metrics for each individual dataset.


\begin{table*}[ht]
\centering
\scalebox{0.70}{
\begin{tabular}{cccccccccc}
    \toprule
    \multirow{2}{*}{\textbf{Method}} & \textbf{Composite} & \multicolumn{3}{c}{\textbf{Faithfulness}} & \multicolumn{3}{c}{\textbf{Plausibility}} & \multicolumn{2}{c}{\textbf{Performance}} \\
    \cmidrule(lr){2-2} \cmidrule(lr){3-5} \cmidrule(lr){6-8} \cmidrule(lr){9-10}
    & NRG ($\uparrow$) & NRG ($\uparrow$) & Comp ($\uparrow$) & Suff ($\downarrow$) & NRG ($\uparrow$) & AUPRC ($\uparrow$) & TF1 ($\uparrow$) & NRG ($\uparrow$) & Acc ($\uparrow$) \\
    
    \midrule
    
    {AA (Grad)} & 0.488 & 0.337 & 0.142~($\pm$0.010) & 0.256~($\pm$0.006) & 0.192 & 58.86~($\pm$3.65) & 27.40~($\pm$0.00) & 0.935 & 93.81~($\pm$0.55) \\
    
    {AA (Input*Grad)} & 0.420 & 0.107 & 0.078~($\pm$0.013) & 0.342~($\pm$0.014) & 0.218 &  44.16~($\pm$1.43) & 45.02~($\pm$0.39) & 0.935 & 93.81~($\pm$0.55) \\
    
    {AA (DeepLIFT)} & 0.453 & 0.122 & 0.085~($\pm$0.006) & 0.340~($\pm$0.018) & 0.302 & 46.50~($\pm$1.32) & 50.18~($\pm$0.32) & 0.935 & 93.81~($\pm$0.55) \\
    
    {AA (IG)} & 0.526 & 0.297 & 0.119~($\pm$0.009) & 0.258~($\pm$0.031) & 0.347 & 49.94~($\pm$1.77) & 50.75~($\pm$0.54) & 0.935 & 93.81~($\pm$0.55) \\
    
    {L2E} & 0.557 & 0.487 & 0.012~($\pm$0.004) & 0.009~($\pm$0.024) & 0.250 & 44.84~($\pm$0.32) & 47.24~($\pm$0.87) & 0.935 & 93.81~($\pm$0.55) \\
    
    {SGT} & 0.632 & 0.555 & 0.147~($\pm$0.024) & 0.113~($\pm$0.031) & 0.371 & 51.38~($\pm$2.47) & 51.35~($\pm$1.64) & 0.971 & 94.40~($\pm$0.57) \\
    
    {FRESH} & 0.330 & 0.837 &  0.219~($\pm$0.057) & 0.000~($\pm$0.000) & 0.152 & 42.06~($\pm$8.84) & 41.19~($\pm$4.01) & 0.000 & 78.78~($\pm$6.48) \\
    
    {A2R} & 0.479 & 0.941 & 0.283~($\pm$0.104) & 0.000~($\pm$0.000) & 0.457 & 63.36~($\pm$6.01) & 46.74~($\pm$6.65) & 0.038 & 79.39~($\pm$11.67) \\
    
    {\methodsp (AA-F)} & 0.639 & 0.706 & 0.292~($\pm$0.051) & 0.171~($\pm$0.038) & 0.329 & 48.13~($\pm$1.14) & 50.96~($\pm$0.93) & 0.882 & 92.97~($\pm$0.44) \\
    
    \midrule
    
    {SGT+P} & 0.596 & 0.507 & 0.139~($\pm$0.032) & 0.137~($\pm$0.026) & 0.355 & 50.38~($\pm$1.45) & 50.98~($\pm$0.46) & 0.928 & 93.70~($\pm$0.88) \\
    
    {FRESH+P} & 0.582 & 0.765 & 0.175~($\pm$0.043) & 0.000~($\pm$0.000) & 0.970 & 84.35~($\pm$0.87) & 71.54~($\pm$0.53) & 0.011 & 78.95~($\pm$5.18) \\
    
    {A2R+P} & 0.695 & 0.953 & 0.290~($\pm$0.016) & 0.000~($\pm$0.000) & 0.978 & 85.56~($\pm$1.01) & 70.97~($\pm$1.03) & 0.154 & 81.26~($\pm$0.52) \\
    
    {\methodsp (DLM-P)} & 0.770 & 0.339 & 0.142~($\pm$0.008) & 0.255~($\pm$0.007) & 0.970 & 84.35~($\pm$0.87) & 71.54~($\pm$0.53) & 1.000 & 94.86~($\pm$0.41) \\
    
    {\methodsp (AA-FP)} & 0.636 & 0.339 & 0.296~($\pm$0.067) & 0.185~($\pm$0.048) & 0.315 & 47.60~($\pm$2.44) & 50.23~($\pm$2.26) & 0.900 & 93.25~($\pm$0.45) \\
    
    {\methodsp (DLM-FP)} & 0.897 & 0.756 & 0.319~($\pm$0.090) & 0.167~($\pm$0.036) & 1.000 & 85.80~($\pm$0.74) & 72.76~($\pm$0.19) & 0.935 & 93.81~($\pm$0.54) \\
    
    {\methodsp (SLM-FP)} & 0.891 & 0.807 & 0.302~($\pm$0.039) & 0.113~($\pm$0.013) & 0.940 & 82.55~($\pm$0.84) & 70.65~($\pm$0.44) & 0.927 & 93.68~($\pm$0.67) \\
    
    \bottomrule 
\end{tabular}
}
\caption{\small{\textbf{Main Results on SST}}}
\label{tab:main:sst}
\end{table*}

\begin{table*}[ht]
\centering
\scalebox{0.70}{
\begin{tabular}{cccccccccc}
    \toprule
    \multirow{2}{*}{\textbf{Method}} & \textbf{Composite} & \multicolumn{3}{c}{\textbf{Faithfulness}} & \multicolumn{3}{c}{\textbf{Plausibility}} & \multicolumn{2}{c}{\textbf{Performance}} \\
    \cmidrule(lr){2-2} \cmidrule(lr){3-5} \cmidrule(lr){6-8} \cmidrule(lr){9-10}
    & NRG ($\uparrow$) & NRG ($\uparrow$) & Comp ($\uparrow$) & Suff ($\downarrow$) & NRG ($\uparrow$) & AUPRC ($\uparrow$) & TF1 ($\uparrow$) & NRG ($\uparrow$) & F1 ($\uparrow$) \\
    
    \midrule
    
    {AA (Grad)} & 0.481 & 0.457 & 0.184~($\pm$0.023) & 0.107~($\pm$0.017) & 0.028 & 13.31~($\pm$0.91) & 5.02~($\pm$0.00) & 0.957 & 95.33~($\pm$0.65) \\
    
    {AA (Input*Grad)} & 0.503 & 0.359 & 0.148~($\pm$0.031) & 0.137~($\pm$0.019) & 0.194 & 8.68~($\pm$0.37) & 37.58~($\pm$0.55) & 0.957 & 95.33~($\pm$0.65) \\
    
    {AA (DeepLIFT)} & 0.468 & 0.259 & 0.122~($\pm$0.029) & 0.172~($\pm$0.022) & 0.187 & 9.00~($\pm$0.16) & 36.15~($\pm$1.45) & 0.957 & 95.33~($\pm$0.65) \\
    
    {AA (IG)} & 0.439 & 0.173 & 0.134~($\pm$0.016) & 0.219~($\pm$0.044) & 0.188 & 8.88~($\pm$0.21) & 36.39~($\pm$1.29) & 0.957 & 95.33~($\pm$0.65) \\
    
    {L2E} & 0.550 & 0.445 & 0.000~($\pm$0.007) & 0.026~($\pm$0.015) & 0.248 & 16.68~($\pm$10.20) & 38.92~($\pm$4.07) & 0.957 & 95.33~($\pm$0.65) \\
    
    {SGT} & 0.553 & 0.474 & 0.124~($\pm$0.053) & 0.071~($\pm$0.064) & 0.184 & 10.05~($\pm$1.23) & 34.64~($\pm$1.67) & 1.000 & 96.33~($\pm$0.76) \\
    
    {FRESH} & 0.645 & 0.732 & 0.234~($\pm$0.034) & 0.000~($\pm$0.000) & 0.305 & 17.02~($\pm$6.22) & 48.26~($\pm$5.87) & 0.899 & 94.00~($\pm$1.44) \\
    
    {A2R} & 0.431 & 0.764 & 0.267~($\pm$0.050) & 0.000~($\pm$0.000) & 0.244 & 35.44~($\pm$21.69) & 19.78~($\pm$25.56) & 0.284 & 79.78~($\pm$7.14) \\
    
    {\methodsp (AA-F)} & 0.601 & 0.744 & 0.505~($\pm$0.134) & 0.122~($\pm$0.100) & 0.189 & 9.14~($\pm$2.51) & 36.28~($\pm$1.84) & 0.870 & 93.33~($\pm$1.61) \\
    
    \midrule
    
    {SGT+P} & 0.586 & 0.604 & 0.152~($\pm$0.013) & 0.022~($\pm$0.004) & 0.183 & 9.16~($\pm$1.59) & 35.33~($\pm$0.41) & 0.971 & 95.66~($\pm$1.16) \\
    
    {FRESH+P} & 0.587 & 0.691 & 0.193~($\pm$0.062) & 0.000~($\pm$0.000) & 1.000 & 94.32~($\pm$0.12) & 89.53~($\pm$1.63) & 0.070 & 74.84~($\pm$12.22) \\
    
    {A2R+P} & 0.585 & 0.764 & 0.267~($\pm$0.076) & 0.000~($\pm$0.000) & 0.991 & 93.53~($\pm$0.93) & 88.77~($\pm$1.22) & 0.000 & 73.22~($\pm$0.75) \\
    
    {\methodsp (DLM-P)} & 0.667 & 0.024 & 0.024~($\pm$0.003) & 0.238~($\pm$0.004) & 1.000 & 94.32~($\pm$0.12) & 89.53~($\pm$1.63) & 0.978 & 95.83~($\pm$0.29) \\
    
    {\methodsp (AA-FP)} & 0.543 & 0.514 & 0.428~($\pm$0.174) & 0.195~($\pm$0.105) & 0.193 & 8.53~($\pm$0.46) & 37.71~($\pm$3.12) & 0.921 & 94.50~($\pm$1.00) \\
    
    {\methodsp (DLM-FP)} & 0.744 & 0.326 & 0.283~($\pm$0.217) & 0.216~($\pm$0.005) & 0.991 & 93.65~($\pm$0.36) & 88.68~($\pm$2.29) & 0.913 & 94.33~($\pm$1.61) \\
    
    {\methodsp (SLM-FP)} & 0.754 & 0.362 & 0.313~($\pm$0.059) & 0.213~($\pm$0.014) & 0.965 & 91.70~($\pm$1.84) &  86.17~($\pm$1.20) & 0.935 & 94.83~($\pm$0.76) \\
    
    \bottomrule 
\end{tabular}
}
\caption{\small{\textbf{Main Results on Movies}}}
\label{tab:main:movies}
\end{table*}

\begin{table*}[ht]
\centering
\scalebox{0.70}{
\begin{tabular}{cccccccccc}
    \toprule
    \multirow{2}{*}{\textbf{Method}} & \textbf{Composite} & \multicolumn{3}{c}{\textbf{Faithfulness}} & \multicolumn{3}{c}{\textbf{Plausibility}} & \multicolumn{2}{c}{\textbf{Performance}} \\
    \cmidrule(lr){2-2} \cmidrule(lr){3-5} \cmidrule(lr){6-8} \cmidrule(lr){9-10}
    & NRG ($\uparrow$) & NRG ($\uparrow$) & Comp ($\uparrow$) & Suff ($\downarrow$) & NRG ($\uparrow$) & AUPRC ($\uparrow$) & TF1 ($\uparrow$) & NRG ($\uparrow$) & Acc ($\uparrow$) \\
    
    \midrule
    
    {AA (Grad)} & 0.537 & 0.504 & 0.331~($\pm$0.012) & 0.352~($\pm$0.007) & 0.130 & 37.33~($\pm$0.62) & 22.65~($\pm$0.00) & 0.977 & 63.56~($\pm$1.27) \\
    
    {AA (Input*Grad)} & 0.573 & 0.361 & 0.249~($\pm$0.018) & 0.385~($\pm$0.008) & 0.383 & 39.56~($\pm$0.54) & 44.43~($\pm$0.40) & 0.977 & 63.56~($\pm$1.27) \\
    
    {AA (DeepLIFT)} & 0.605 & 0.346 & 0.254~($\pm$0.035) & 0.403~($\pm$0.042) & 0.491 & 42.82~($\pm$1.83) & 51.72~($\pm$1.26) & 0.977 & 63.56~($\pm$1.27) \\
    
    {AA (IG)} & 0.578 & 0.327 & 0.216~($\pm$0.007) & 0.378~($\pm$0.010) & 0.429 & 40.07~($\pm$5.47) & 48.34~($\pm$3.16) & 0.977 & 63.56~($\pm$1.27) \\
    
    {L2E} & 0.544 & 0.493 & 0.005~($\pm$0.003) & 0.010~($\pm$0.008) & 0.161 & 23.56~($\pm$1.09) & 37.80~($\pm$1.10) & 0.977 & 63.56~($\pm$1.27) \\
    
    {SGT} & 0.618 & 0.367 & 0.197~($\pm$0.040) & 0.324~($\pm$0.015) & 0.491 & 43.68~($\pm$4.68) & 51.00~($\pm$3.05) & 0.995 & 64.35~($\pm$0.46) \\
    
    {FRESH} & 0.302 & 0.546 & 0.037~($\pm$0.036) & 0.000~($\pm$0.000) & 0.261 & 32.35~($\pm$7.66) & 39.37~($\pm$0.70) & 0.101 & 24.81~($\pm$3.46) \\
    
    {A2R} & 0.277 & 0.516 & 0.014~($\pm$0.021) & 0.000~($\pm$0.000) & 0.282 & 41.61~($\pm$3.85) & 33.12~($\pm$9.06) & 0.032 & 21.77~($\pm$1.31) \\
    
    {\methodsp (AA-F)} & 0.690 & 0.538 & 0.297~($\pm$0.141) & 0.286~($\pm$0.084) & 0.554 & 46.97~($\pm$3.41) & 53.99~($\pm$1.66) & 0.978 & 63.58~($\pm$0.61) \\
    
    \midrule
    
    {SGT+P} & 0.601 & 0.367 & 0.201~($\pm$0.032) & 0.328~($\pm$0.022) & 0.436 & 41.30~($\pm$6.70) & 47.95~($\pm$1.65) & 1.000 & 64.57~($\pm$0.33) \\
    
    {FRESH+P} & 0.504 & 0.515 & 0.013~($\pm$0.021) & 0.013~($\pm$0.021) & 0.997 & 76.07~($\pm$1.63) & 69.76~($\pm$0.27) & 0.000 & 20.36~($\pm$0.66) \\
    
    {A2R+P} & 0.488 & 0.500 & 0.001~($\pm$0.001) & 0.000~($\pm$0.000) & 0.951 & 73.59~($\pm$0.81) & 67.63~($\pm$1.54) & 0.012 & 20.91~($\pm$0.48) \\
    
    {\methodsp (DLM-P)} & 0.751 & 0.267 & 0.180~($\pm$0.016) & 0.390~($\pm$0.035) & 0.997 & 76.07~($\pm$1.63) & 69.76~($\pm$0.27) & 0.990 & 64.13~($\pm$0.46) \\
    
    {\methodsp (AA-FP)} & 0.685 & 0.551 & 0.395~($\pm$0.109) & 0.381~($\pm$0.101) & 0.537 & 45.21~($\pm$4.46) & 53.91~($\pm$3.23) & 0.968 & 63.14~($\pm$0.33) \\
    
    {\methodsp (DLM-FP)} & 0.814 & 0.492 & 0.293~($\pm$0.043) & 0.321~($\pm$0.070) & 0.997 & 76.38~($\pm$0.57) & 69.52~($\pm$0.24) & 0.953 & 62.50~($\pm$1.34) \\
    
    {\methodsp (SLM-FP)} & 0.807 & 0.494 & 0.390~($\pm$0.087) & 0.424~($\pm$0.110) & 0.983 & 75.12~($\pm$0.41) & 69.25~($\pm$0.41) & 0.944 & 62.09~($\pm$2.12) \\
    
    \bottomrule
\end{tabular}
}
\caption{\small{\textbf{Main Results on CoS-E}}}
\label{tab:main:cose}
\end{table*}

\begin{table*}[ht]
\centering
\scalebox{0.70}{
\begin{tabular}{cccccccccc}
    \toprule
    \multirow{2}{*}{\textbf{Method}} & \textbf{Composite} & \multicolumn{3}{c}{\textbf{Faithfulness}} & \multicolumn{3}{c}{\textbf{Plausibility}} & \multicolumn{2}{c}{\textbf{Performance}} \\
    \cmidrule(lr){2-2} \cmidrule(lr){3-5} \cmidrule(lr){6-8} \cmidrule(lr){9-10}
    & NRG ($\uparrow$) & NRG ($\uparrow$) & Comp ($\uparrow$) & Suff ($\downarrow$) & NRG ($\uparrow$) & AUPRC ($\uparrow$) & TF1 ($\uparrow$) & NRG ($\uparrow$) & F1 ($\uparrow$) \\
    
    \midrule
    
    {AA (Grad)} & 0.498 & 0.462 & 0.222~($\pm$0.028) & 0.120~($\pm$0.018) & 0.035 & 22.27~($\pm$0.17) & 13.81~($\pm$0.00) & 0.997 & 69.80~($\pm$0.60) \\
    
    {AA (Input*Grad)} & 0.506 & 0.289 & 0.225~($\pm$0.048) & 0.260~($\pm$0.059) & 0.231 & 18.51~($\pm$0.23) & 43.45~($\pm$0.05) & 0.997 & 69.80~($\pm$0.60) \\
    
    {AA (DeepLIFT)} & 0.493 & 0.249 & 0.225~($\pm$0.012) & 0.292~($\pm$0.014) & 0.234 & 18.80~($\pm$0.19) & 43.51~($\pm$0.04) & 0.997 & 69.80~($\pm$0.60) \\
    
    {AA (IG)} & 0.499 & 0.280 & 0.162~($\pm$0.086) & 0.222~($\pm$0.086) & 0.220 & 18.71~($\pm$0.40) & 41.79~($\pm$1.33) & 0.997 & 69.80~($\pm$0.60) \\
    
    {L2E} & 0.522 & 0.366 & 0.007~($\pm$0.006) & 0.042~($\pm$0.024) & 0.205 & 24.48~($\pm$2.71) & 32.63~($\pm$6.12) & 0.997 & 69.80~($\pm$0.60) \\
    
    {SGT} & 0.594 & 0.564 & 0.214~($\pm$0.105) & 0.033~($\pm$0.077) & 0.224 & 18.60~($\pm$0.42) & 42.42~($\pm$0.51) & 0.995 & 69.73~($\pm$0.13) \\
    
    {FRESH} & 0.675 & 0.571 & 0.176~($\pm$0.029) & 0.000~($\pm$0.000) & 0.617 & 24.68~($\pm$7.98) & 48.02~($\pm$3.04) & 0.838 & 64.47~($\pm$3.41) \\
    
    {A2R} & 0.217 & 0.404 & -0.010~($\pm$0.029) & 0.000~($\pm$0.000) & 0.249 & 18.72~($\pm$0.67) & 45.45~($\pm$0.02) & 0.000 & 36.39~($\pm$0.00) \\
    
    {\methodsp (AA-F)} & 0.711 & 0.956 & 0.505~($\pm$0.050) & -0.071~($\pm$0.020) & 0.236 & 18.82~($\pm$0.40) & 43.68~($\pm$0.38) & 0.939 & 66.17~($\pm$4.58) \\
    
    \midrule
    
    {SGT+P} & 0.630 & 0.665 & 0.280~($\pm$0.029) & 0.283~($\pm$0.039) & 0.226 & 18.63~($\pm$0.52) & 42.71~($\pm$0.39) & 1.000 & 69.91~($\pm$0.81) \\
    
    {FRESH+P} & 0.491 & 0.413 & 0.000~($\pm$0.013) & 0.000~($\pm$0.000) & 0.999 & 71.80~($\pm$0.27) & 77.94~($\pm$0.57) & 0.060 & 38.41~($\pm$5.34) \\
    
    {A2R+P} & 0.516 & 0.422 & 0.011~($\pm$0.024) & 0.000~($\pm$0.000) & 0.977 & 70.86~($\pm$1.30) & 76.21~($\pm$1.68) & 0.150 & 41.42~($\pm$8.73) \\
    
    {\methodsp (DLM-P)} & 0.708 & 0.123 & 0.127~($\pm$0.010) & 0.322~($\pm$0.017) & 0.999 & 71.80~($\pm$0.27) & 77.94~($\pm$0.57) & 1.000 & 69.91~($\pm$0.76) \\
    
    {\methodsp (AA-FP)} & 0.706 & 1.000 & 0.545~($\pm$0.045) & -0.077~($\pm$0.099) & 0.231 & 19.13~($\pm$0.71) & 42.66~($\pm$1.18) & 0.888 & 66.17~($\pm$4.58) \\
    
    {\methodsp (DLM-FP)} & 0.751 & 0.327 & 0.135~($\pm$0.072) & 0.165~($\pm$0.029) & 0.998 & 71.89~($\pm$0.41) & 77.63~($\pm$0.62) & 0.929 & 67.53~($\pm$1.06) \\
    
    {\methodsp (SLM-FP)} & 0.784 & 0.377 & 0.198~($\pm$0.038) & 0.171~($\pm$0.027) & 0.997 & 71.69~($\pm$0.21) & 77.79~($\pm$0.09) & 0.979 & 69.20~($\pm$1.58) \\
    
    \bottomrule 
\end{tabular}
}
\caption{\small{\textbf{Main Results on MultiRC}}}
\label{tab:main:multirc}
\end{table*}

\begin{table*}[ht]
\centering
\scalebox{0.70}{
\begin{tabular}{cccccccccc}
    \toprule
    \multirow{2}{*}{\textbf{Method}} & \textbf{Composite} & \multicolumn{3}{c}{\textbf{Faithfulness}} & \multicolumn{3}{c}{\textbf{Plausibility}} & \multicolumn{2}{c}{\textbf{Performance}} \\
    \cmidrule(lr){2-2} \cmidrule(lr){3-5} \cmidrule(lr){6-8} \cmidrule(lr){9-10}
    & NRG ($\uparrow$) & NRG ($\uparrow$) & Comp ($\uparrow$) & Suff ($\downarrow$) & NRG ($\uparrow$) & AUPRC ($\uparrow$) & TF1 ($\uparrow$) & NRG ($\uparrow$) & F1 ($\uparrow$) \\
    
    \midrule
    
    {AA (Grad)} & 0.587 & 0.518 & 0.313~($\pm$0.009) & 0.380~($\pm$0.025) & 0.244 & 59.80~($\pm$1.32) & 15.27~($\pm$0.00) & 0.999 & 90.78~($\pm$0.27) \\
    
    {AA (Input*Grad)} & 0.503 & 0.287 & 0.205~($\pm$0.005) & 0.446~($\pm$0.020) & 0.223 & 32.98~($\pm$1.37) & 43.13~($\pm$0.86) & 0.999 & 90.78~($\pm$0.27) \\
    
    {AA (DeepLIFT)} & 0.508 & 0.270 & 0.195~($\pm$0.012) & 0.448~($\pm$0.014) & 0.254 & 33.47~($\pm$1.31) & 46.44~($\pm$0.04) & 0.999 & 90.78~($\pm$0.27) \\
    
    {AA (IG)} & 0.596 & 0.473 & 0.308~($\pm$0.011) & 0.414~($\pm$0.020) & 0.317 & 47.83~($\pm$1.04) & 37.87~($\pm$1.39) & 0.999 & 90.78~($\pm$0.27) \\
    
    {L2E} & 0.606 & 0.460 & 0.009~($\pm$0.015) & 0.036~($\pm$0.022) & 0.358 & 58.11~($\pm$0.97) & 31.35~($\pm$0.27) & 0.999 & 90.78~($\pm$0.27) \\
    
    {SGT} & 0.595 & 0.503 & 0.288~($\pm$0.025) & 0.361~($\pm$0.038) & 0.298 & 42.46~($\pm$3.03) & 41.70~($\pm$1.78) & 0.985 & 90.23~($\pm$0.16) \\
    
    {FRESH} & 0.518 & 0.661 & 0.120~($\pm$0.075) & 0.000~($\pm$0.000) & 0.361 & 38.77~($\pm$6.82) & 53.71~($\pm$3.30) & 0.530 & 72.92~($\pm$8.71) \\
    
    {A2R} & 0.273 & 0.564 & 0.053~($\pm$0.048) & 0.000~($\pm$0.000) & 0.256 & 48.48~($\pm$11.14) & 29.54~($\pm$24.72) & 0.000 & 52.72~($\pm$14.08) \\
    
    {\methodsp (AA-F)} & 0.622 & 0.539 & 0.330~($\pm$0.018) & 0.383~($\pm$0.055) & 0.340 & 45.29~($\pm$3.02) & 43.69~($\pm$1.98) & 0.987 & 90.31~($\pm$0.19) \\
    
    \midrule
    
    {SGT+P} & 0.608 & 0.524 & 0.286~($\pm$0.034) & 0.339~($\pm$0.032) & 0.311 & 43.03~($\pm$1.69) & 42.59~($\pm$1.63) & 0.988 & 90.36~($\pm$0.08) \\
    
    {FRESH+P} & 0.746 & 0.695 & 0.143~($\pm$0.072) & 0.000~($\pm$0.000) & 1.000 & 87.85~($\pm$0.13) & 77.63~($\pm$0.35) & 0.544 & 73.44~($\pm$12.88) \\
    
    {A2R+P} & 0.800 & 0.751 & 0.182~($\pm$0.097) & 0.000~($\pm$0.000) & 0.992 & 87.30~($\pm$0.44) & 77.31~($\pm$0.72) & 0.656 & 77.31~($\pm$0.72) \\
    
    {\methodsp (DLM-P)} & 0.842 & 0.525 & 0.311~($\pm$0.011) & 0.371~($\pm$0.032) & 1.000 & 87.85~($\pm$0.13) & 77.63~($\pm$0.35) & 1.000 & 90.80~($\pm$0.33) \\
    
    {\methodsp (AA-FP)} & 0.626 & 0.529 & 0.341~($\pm$0.008) & 0.406~($\pm$0.046) & 0.363 & 44.79~($\pm$0.81) & 47.18~($\pm$0.83) & 0.985 & 90.21~($\pm$0.08) \\
    
    {\methodsp (DLM-FP)} & 0.857 & 0.588 & 0.335~($\pm$0.018) & 0.346~($\pm$0.023) & 0.991 & 86.99~($\pm$0.40) & 77.53~($\pm$0.15) & 0.992 & 90.51~($\pm$0.12) \\
    
    {\methodsp (SLM-FP)} & 0.864 & 0.603 & 0.353~($\pm$0.017) & 0.356~($\pm$0.015) & 0.994 & 87.58~($\pm$0.14) & 77.22~($\pm$0.28) & 0.994 & 90.59~($\pm$0.09) \\
    
    \bottomrule 
\end{tabular}
}
\caption{\small{\textbf{Main Results on e-SNLI}}}
\label{tab:main:esnli}
\end{table*}

\end{document}